%% file: main.tex
\documentclass{article}
\usepackage[final,nonatbib]{neurips_2022}

\usepackage[utf8]{inputenc} 
\usepackage[T1]{fontenc}    
\usepackage{hyperref}       
\usepackage{url}            
\usepackage{booktabs}       
\usepackage{amsfonts}       
\usepackage{nicefrac}       
\usepackage{microtype}      
\usepackage{xcolor}         
\usepackage{graphicx}
\usepackage{multirow}
\usepackage{makecell}
\usepackage{tikz}
\usepackage[hypcap=false]{caption}\usepackage{tabularx}
\usepackage[export]{adjustbox}

\usepackage{multirow}
\usepackage{subcaption}

\usepackage{amsmath,amssymb} 
\usepackage{color}
\usepackage{xspace}

\usepackage{cleveref}

\usepackage{wrapfig}
\usepackage{array}
\newcommand{\PreserveBackslash}[1]{\let\temp=\\#1\let\\=\temp}
\newcolumntype{C}[1]{>{\PreserveBackslash\centering}p{#1}}
\newcolumntype{R}[1]{>{\PreserveBackslash\raggedleft}p{#1}}
\newcolumntype{L}[1]{>{\PreserveBackslash\raggedright}p{#1}}
\usepackage{titlesec}
\titlespacing*{\subsection}
  {0pt}{0.2\baselineskip}{0.2\baselineskip}
  \titlespacing*{\section}
  {0pt}{0.3\baselineskip}{0.3\baselineskip}

\input{macros}

\title{Quo Vadis: Is Trajectory Forecasting the Key Towards Long-Term Multi-Object Tracking?}

\author{%
Patrick Dendorfer
 \quad 
 Vladimir Yugay 
 \quad 
 Aljoša Ošep 
 \quad 
 Laura Leal-Taix\'e
     \\
     \\
    Technical University of Munich
    \\
    \\
    \texttt{\{patrick.dendorfer, vladimir.yugay, aljosa.osep, leal.taixe\}@tum.de}
}
\begin{document}
\maketitle
\begin{abstract}
Recent developments in monocular multi-object tracking have been very successful in tracking visible objects and bridging short occlusion gaps, mainly relying on data-driven appearance models. 
While we have significantly advanced short-term tracking performance, bridging longer occlusion gaps remains elusive: state-of-the-art object trackers only bridge less than $10\%$ of occlusions longer than three seconds. 
We suggest that the missing key is reasoning about future trajectories over a longer time horizon. Intuitively, the longer the occlusion gap, the larger the search space for possible associations. %
In this paper, we show that even a small yet diverse set of trajectory predictions for moving agents will significantly reduce this search space and thus improve long-term tracking robustness. Our experiments suggest that the crucial components of our approach are reasoning in a bird's-eye view space and generating a small yet diverse set of forecasts while accounting for their localization uncertainty. This way, we can advance state-of-the-art trackers on the \MOTChallenge dataset and significantly improve their long-term tracking performance. This paper's source code and experimental data are available at \url{https://github.com/dendorferpatrick/QuoVadis}.
\end{abstract}

\section{Introduction}
Multi-object tracking (MOT) is a long-standing research problem with applications ranging from real-time dynamic situational awareness for robot navigation~\cite{Geiger12CVPR,dendorfer20ijcv,bdd100k,martin2021jrdb,sun20CVPR,Caesar20CVPR}, traffic monitoring~\cite{Wen15arxiv}, studying animal behavior~\cite{pedersen20203d} and monitoring biological phenomena~\cite{anjum2020ctmc}.

\begin{figure}[ht]
\begin{subfigure}[t]{0.45\textwidth}
    \includegraphics[width=\textwidth]{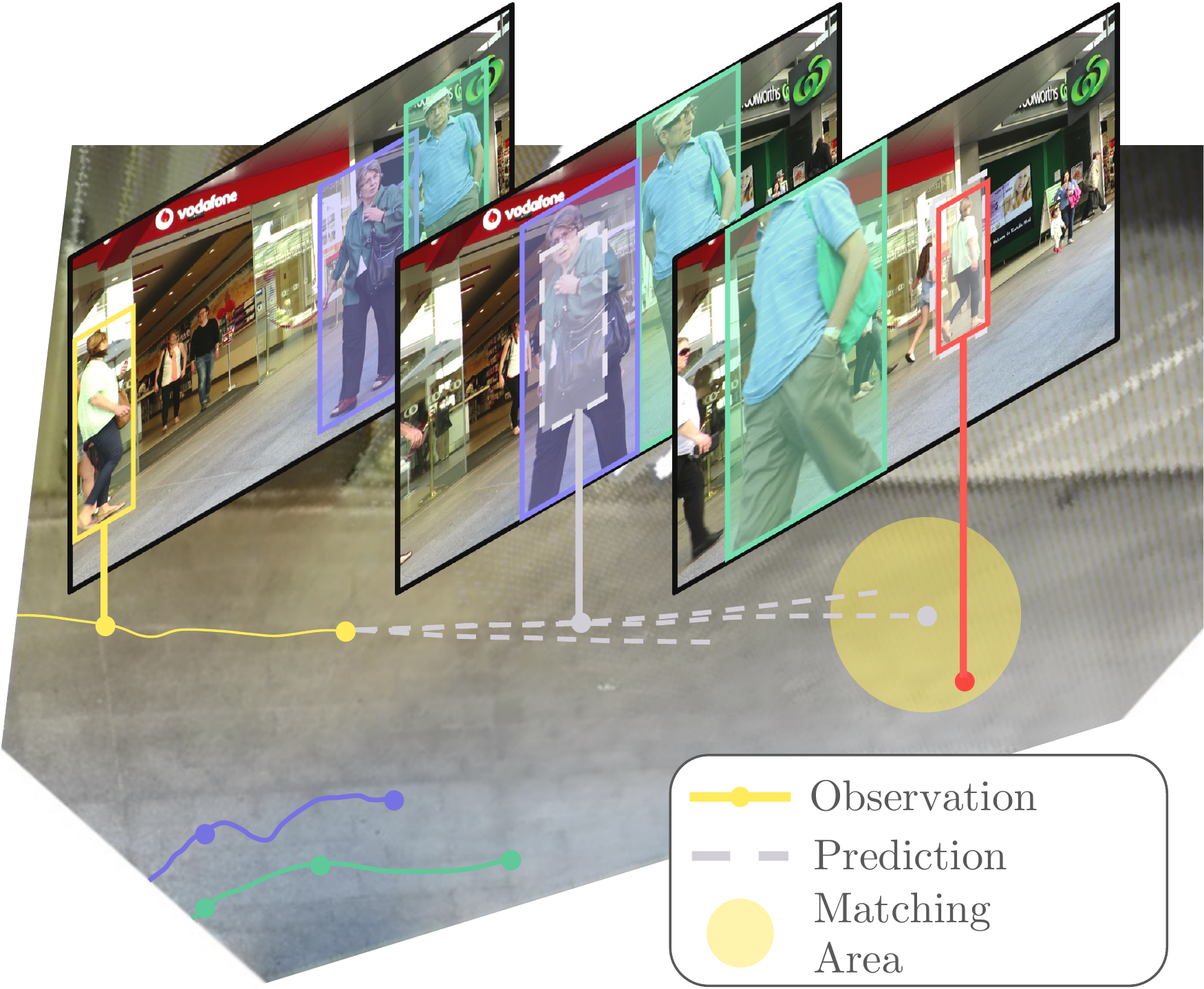}
     \caption{Illustration of our method.}
     \label{figure:teaser:method}
     \end{subfigure}
     \hfill
 \begin{subfigure}[t]{0.5\textwidth}   
    \includegraphics[width=\textwidth]{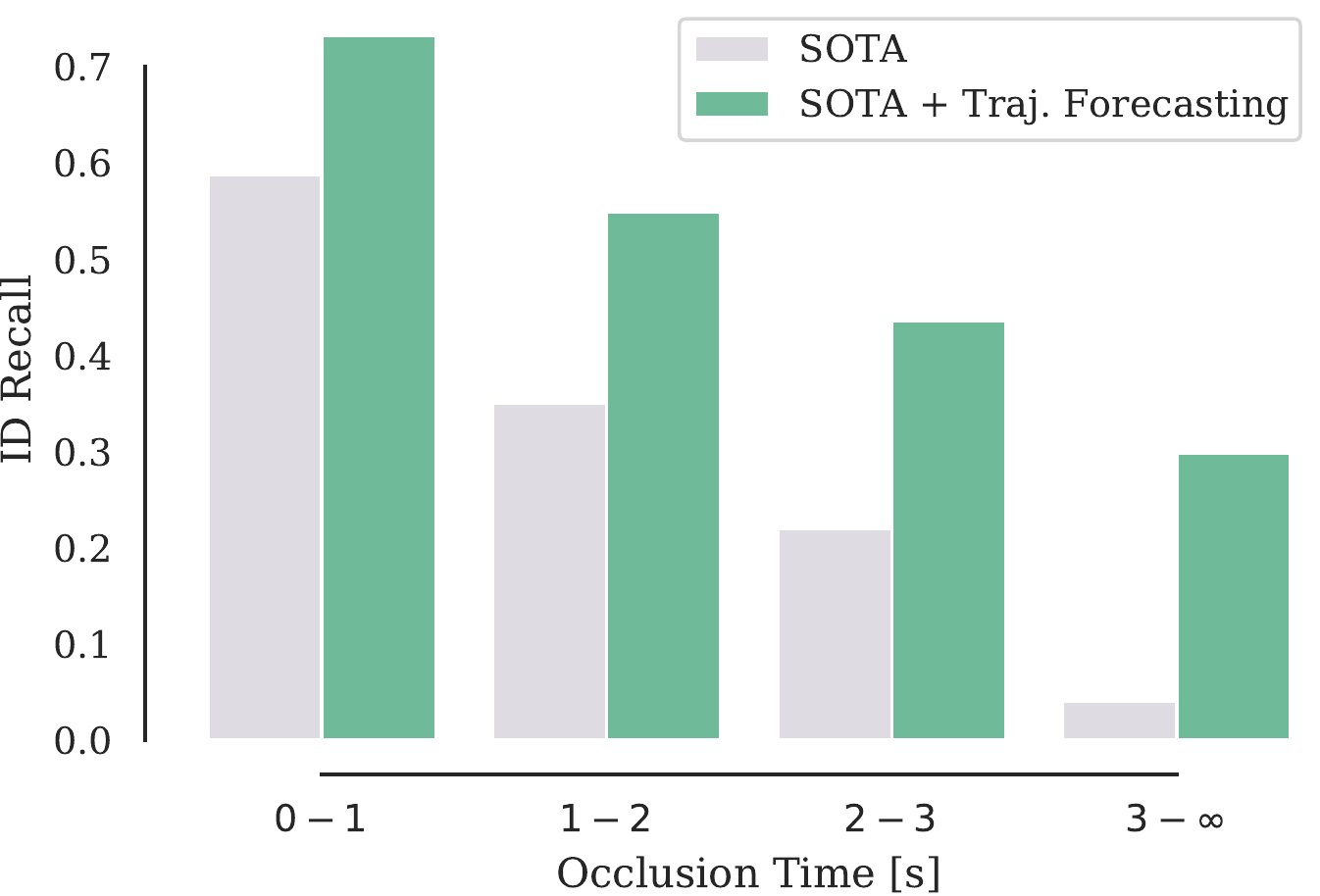}
     \caption{ID recall.}
     \label{figure:teaser:occlusion}
    \end{subfigure}
    \caption{
    State-of-the-art methods for vision-based MOT can successfully track visible objects and bridge short occlusion gaps; however, they fail at long-term tracking. %
    (a) To bridge occlusion gaps, we lift monocular 2D detections to 3D world space, in which we reason about their possible future locations. This transformation allows us to reconnect detections that undergo long occlusions. As \textcolor{darkyellow}{\textbf{yellow track}} becomes occluded, our method predicts a small set of plausible future locations in 3D. In turn, we correctly associate \textcolor{darkred}{\textbf{red detection}} to the \textcolor{darkyellow}{\textbf{yellow track}} by accounting for the forecast uncertainty area.
    (b) As can be seen from the ratio of correct track association after different occlusion time lengths for the \textcolor{darkgrey}{\textbf{prior work}} and \textcolor{darkgreen}{\textbf{our method}}, this approach allows us to significantly improve long-term tracking capabilities and gap longer occlusion gaps. \textit{Best seen in color.}
    }
    \label{figure:teaser}
    \vspace{-0.1cm} 
\end{figure}

State-of-the-art MOT methods~\cite{xu20cvpr,Braso20CVPR,Bergmann19ICCV,zhou2020tracking,xu2021transcenter,tokmakov2021learning} combine regression
\cite{xu20cvpr,Bergmann19ICCV} and combinatorial optimization~\cite{Braso20CVPR} in conjunction with identity re-identification (ReID) models~\cite{LealTaixe16CVPRW,Son17CVPR,Braso20CVPR,xu20cvpr,Voigtlaender19CVPR,Bergmann19ICCV} to track objects in the image space. 
Such approaches have been very successful for tracking visible objects and bridging \textit{short-term} occlusions. However, as can be seen in \Cref{figure:teaser:occlusion}, \textit{long-term tracking} remains an open challenge: state-of-the-art methods successfully bridge $50\%$ of occlusions within one second, falling below $10\%$ when the occlusion extends for more than 3 seconds. This is often not reflected in standard benchmarks~\cite{dendorfer20ijcv,Geiger12CVPR,Wen15arxiv,bdd100k}, as long-term occlusions are statistically rare. 

In the past, combining ReID models with simple motion models has been immensely helpful~\cite{dendorfer20ijcv} for short-term tracking. 
Nonetheless, as the occlusion time becomes longer, the set of possible associations grows exponentially with the increasing gap length~\cite{Reid79TAC}. This combinatorial complexity hinders the ability of visual-based ReID models to disambiguate between objects. Consequently, we believe that ReID models are insufficient to resolve long-term occlusions. However, continued efforts to develop stronger appearance models will remain an important research direction in vision-based MOT. Tracking moving pedestrians during occlusions is challenging, and simple linear motion models fail since human motion is complex and driven by non-observable factors such as goals, intent, or simply preferences. Therefore, we propose an alternative in this work: using long-term trajectory forecasting in order to prune down the combinatorial search space of feasible trajectory continuations.

As the main contribution of this paper, we carefully study \textit{what is needed} to leverage trajectory forecasting for multi-object tracking, as we have recently witnessed rapid progress in learning-based trajectory forecasting~\cite{stanforddronedataset,liang2020garden,Gupta18CVPR,Alahi16CVPR}. However, these methods operate in a fully-observed, metric bird's-eye view (BEV) space, effectively disentangling the effect of the perspective projection on reasoning about motion. By contrast, monocular MOT methods only observe a projection of the visible portion of our 3D space. Our analysis reveals that we can bridge this gap by localizing trajectories in BEV-space, but crucially, the localization of 2D bounding boxes in BEV must be \textit{temporally coherent}. We achieve this by estimating a single homography per sequence in a data-driven manner. 

Forecasting methods can reason beyond simple linear extrapolations, predict multiple possible future outcomes, and account for social interactions. But are these all necessary ingredients for bridging complex and long-term occlusions? Our study suggests that the key ingredient is to estimate a set of forecasts that can possibly cover several diverging future paths with only a handful of samples and account for prediction uncertainty. 

Our \textit{trajectory forecasting} approach can be applied to improve the long-term tracking capabilities of existing object tracking methods. In particular, by applying our framework on top of the state-of-the-art method \cite{byte_track} of the \MOTChallenge benchmark, we improve the performance on HOTA on \MOTSEVENTEEN by $0.09$pp and \MOTTWENTY by $0.10$pp and further decrease the number of IDSW by $93$ and $36$, respectively. We hope our conclusions will encourage the community to continue investigating how 3D reconstruction and trajectory forecasting improve single-camera long-term tracking. 

We summarize our \textbf{main contributions} as follows:
we (i) present a study on how we can reconcile two related fields of research on vision-based trajectory forecasting and monocular multi-object tracking. Our study reveals that the core component of this interplay is temporally coherent reasoning about motion in 3D space. 
We (ii) utilize a synthetic MOT dataset to study how to localize objects in 3D BEV space in a manner that facilitates robust reasoning about plausible future motion and which are the core forecasting components needed to bridge longer occlusion gaps; 
Finally, (iii) we demonstrate that we can generalize our conclusions from synthetic sandbox to real-world monocular \MOTChallenge sequences and demonstrate that our recipe can be used to improve long-term tracking performance for several object trackers.

\section{Preliminaries}
This section discusses the fundamentals of vision-based multi-object tracking and trajectory forecasting, the current state-of-the-art, and analyzes failure cases.
\subsection{Multi-object Tracking}
\label{sec:problemdefinition}
Monocular multi-object tracking (MOT) is the task of localizing objects as bounding boxes in image sequences and assigning them an identity-preserving unique ID. 
State-of-the-art methods decompose the problem into object detection and detection association. 

\PAR{Quantifying tracking errors.} 
The \textit{detection} aspect of the task is commonly quantified by counting per-frame detection errors over the sequence. 
To quantify \textit{association} errors, we count \textit{identity switches} (IDSW) (\ie, wrong ID swaps or re-initializing a ground-truth track with a different tracking ID) and \textit{identity transfers} (IDTR) (\ie, incorrectly linking two different objects with the same tracking ID). While a successful association over occlusion gaps decreases the number of IDSW, a wrong association between tracklets leads to an IDTR instead.
Recently introduced HOTA~\cite{luiten20ijcv} metric separately evaluates object detection and temporal association aspects of the tracking task. 
Temporal association is quantified via \textit{association accuracy} (AssA) term, that quantifies \textit{association recall} (AssRe) and \textit{association precision} (AssPr). The AssRe term accounts for IDSW errors, while AssPr accounts for IDTR errors. 

\PAR{Are all identity errors created equal?} 
Correct track association of objects that undergo longer occlusion gaps is especially challenging because the appearance and position of an object may drastically change. 
In MOT datasets~\cite{dendorfer20ijcv,Geiger12CVPR} the majority of occlusions are short occlusions ($\leq 2s$). 
Hence, solving rare long occlusions ($>2s$) does not significantly impact the model performance. As a result, long-term tracking is commonly overlooked in the literature. 
This can be seen in \Cref{figure:teaser:occlusion}: state-of-the-art methods bridge less than $10\%$ gaps beyond three-second-long occlusions.

\PAR{Prior work.} Early tracking methods focus on combinatorial optimization~\cite{Zhang08CVPR,Huang08ECCV,Li09CVPR,Wu09IJCV,Milan13CVPR} and hand-crafting visual and motion-based descriptors~\cite{Milan14TPAMI,leal14cvpr,Choi15ICCV}, especially beneficial in the era of unreliable object detectors. 
State-of-the-art methods for monocular visual MOT are data-driven and primarily rely on appearance. Regression-based methods~\cite{Bergmann19ICCV,xu20cvpr,zhou2020tracking} can localize objects even when object detections are missing, often used in conjunction with ReID models to bridge short occlusions. However, regression models fail when an occluded person appears at a distant image position. For solving long-term occlusion, discrete optimization methods, combined with end-to-end learning based on graph neural networks~\cite{Braso20CVPR,weng20CVPR,zaech2022learnable}, construct large graphs stretching over multiple seconds leading to high computational costs and complexity. 
Motion has always played an essential role in visual tracking~\cite{bewley16icip,Geiger14TPAMI,Bergmann19ICCV}, especially beneficial in 3D where it is dis-entangled from projective distortion~\cite{leibe08TPAMI,Osep17ICRA,Hu18arxiv}. 
The interplay between reasoning in 3D for monocular pedestrian tracking and linear motion models was first investigated in~\cite{khurana2021detecting}. 

Identity preservation is important in several applications, ranging from video editing, safety camera analysis, and social robots interacting with humans to autonomous driving. We only have access to a single RGB camera in several application scenarios. Exceptions are autonomous driving datasets~\cite{Geiger12CVPR,Caesar20CVPR} that generally provide 3D sensory data, together with 3D track information. However, only a handful of object tracks contain occlusion gaps longer than $2s$: $0.6\%$ in BDD100K~\cite{bdd100k} and $4\%$ in widely-used KITTI tracking~\cite{Geiger12CVPR} dataset. Therefore, autonomous driving datasets are, at the moment, not well suited for studying long-term tracking. Instead we conduct our experiments and analysis using \MOTChallenge~\cite{dendorfer20ijcv} dataset, where $19.4\%$ of tracks undergo \textit{long}  ($>2s$) occlusions gaps.

We hypothesize that bridging long-term gaps requires understanding the projection geometry and motion models that can reason about plausible diverging future paths and non-linear motion.

\subsection{Trajectory Forecasting}
\label{sec:forecasting_basics}
Pedestrian trajectory forecasting has been studied independently of the closely related task of object tracking. Forecasting is challenging because (i) human behavior and, therefore, future motion underlies the effect of complex social and scene interactions and latent navigating intent. Moreover, (ii) entire scene geometry is usually not directly visible to the observer, and in general, it is difficult to localize past trajectories precisely. 
To this end, existing models use standard datasets \cite{ETH-data, UCY-data, liang2020garden, stanforddronedataset} and study forecasting in idealized conditions: given an accurate bird's-eye view of the scene and perfectly-localized past trajectories to predict trajectory continuations in metric space.

\PAR{Quantifying forecasting accuracy.} Forecasting performance is measured in metric space as $L_2$ distance between the prediction and ground-truth trajectory (as final displacement error, FDE, or average displacement error, ADE) \wrt top-k forecasts (commonly $k=20$). We note that this approach mainly incentivizes high forecasting recall and neglects forecasting precision which is important for the application of forecasting methods~\cite{dendorfer21iccv}. 

\PAR{Prior work.} 
Early forecasting methods were deterministic, firstly based on physical models~\cite{social_force}, and later on data-driven LSTM-based encoder-decoder networks~\cite{Alahi16CVPR} methods, focusing on modeling social~\cite{Gupta18CVPR, Alahi16CVPR, social_ways} and scene~\cite{sadeghian2018sophie, bigat} interactions. The forecasting task is inherently uncertain, and we need to express the stochasticity in the model. With learning a distribution of possible future trajectories, generative models~\cite{Gupta18CVPR, sadeghian2018sophie, bigat, social_ways, dendorfer20accv, dendorfer21iccv} have emerged as state-of-the-art prediction methods. Recent efforts have been explicitly focusing on conditioning forecasting on estimated pedestrian goal/intent~\cite{dendorfer20accv,mangalam2020pecnet,Mangalam_2021_ICCV} and estimating multimodal posterior distributions~\cite{liang2020garden,dendorfer21iccv} that yield diverse trajectories that cover different plausible directions. These deep neural network approaches can model complex and non-linear trajectories beyond simple linear models. 

Can we bring the \textit{two worlds} together, and if so, \textit{how}? Furthermore, which of the aforementioned aspects of forecasting methods (\ie, stochasticity, non-linearity, multimodality, diversity, accounting for interactions) are \textit{important} in the context of multi-object tracking? These are the questions we discuss in the following sections.

\section{Methodology}
\label{sec:method}
In this section, we present our method for long-term multi-object tracking based on trajectory forecasting in bird's-eye view (BEV) scene representation. Simply applying trajectory prediction to multi-object tracking is not trivially possible, as object trajectories observed in the image space break multiple assumptions of real-world trajectory prediction. While trajectory prediction works in bird's-eye view coordinates, the motion and size of objects in image space depend on the camera's intrinsic parameters, orientation, and position. In addition, we face temporal (limited length of observation), association (association errors along with observation), and measurement (imprecise localization of objects) uncertainties of the trajectories. Contrarily, objects are represented as bounding boxes in the image instead of single 2D positions for the object tracking task. To bridge the gap between prediction and tracking, we must find a transformation from the image to the real space. We assume objects move on a planar ground to formulate such a transformation. Thus, the bottom-center points of detection bounding boxes $p$ can be mapped to a 2D BEV coordinate $x$ via an initially unknown homography transformation $H$ that relates the homogeneous coordinates as $x \propto H \cdot p $. 

\begin{figure}[t]
    \centering
    \includegraphics[width=\textwidth]{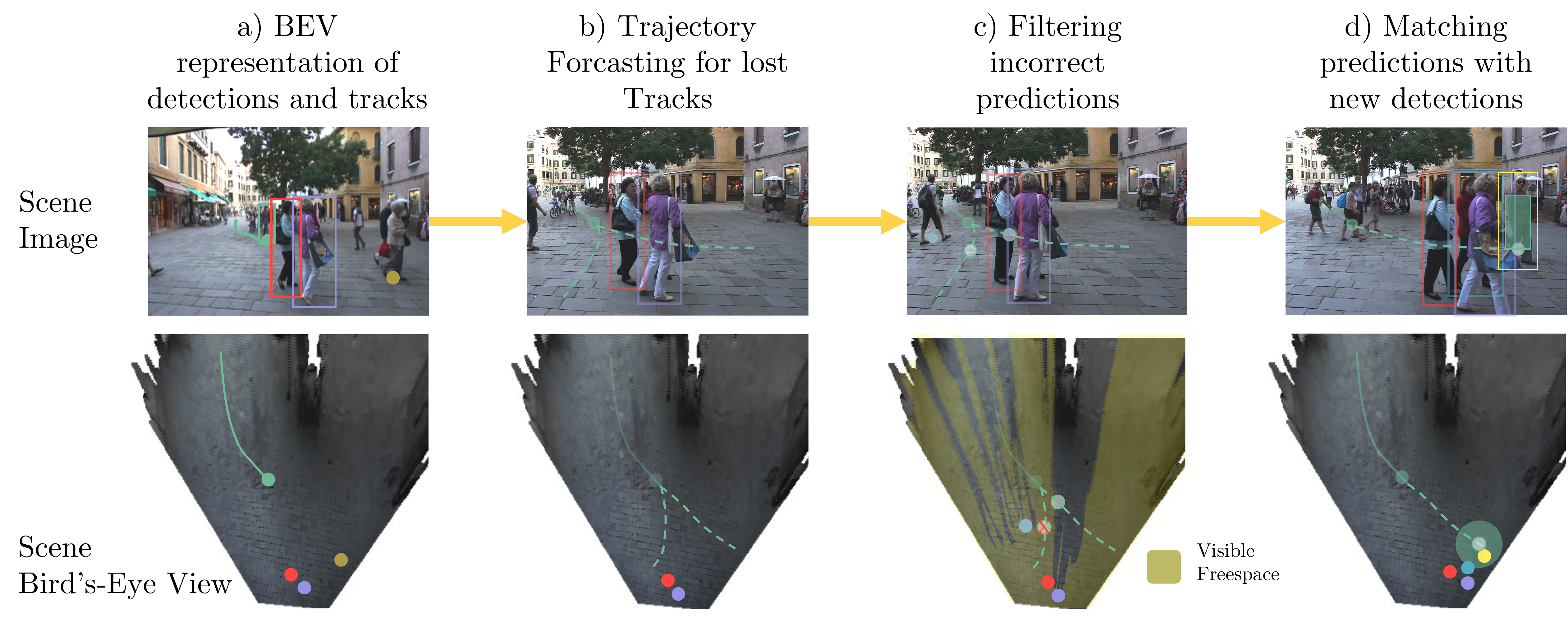} 
    \caption{\textbf{Our method:} we bridge long-term occlusions by (a) localizing object tracks in BEV via estimated homography and (b) forecasting future trajectories for \textit{lost} tracks. We (d) continually aim to match these \textit{inactive} track predictions with new object detections and remove incorrect predictions under a visibility constraint (c).}
    \label{figure:method:Overview}
\end{figure}

\PAR{Overview.}
Given a monocular video sequence captured from a stationary camera from \textit{arbitrary} viewpoint, we first estimate the homography $H$, which maps the image plane to the 3D world ground-plane for the whole sequence (\Cref{sec:method:homography}). Then, we incorporate our model into an online tracker that takes a monocular tracker output and localizes tracks and detections in BEV space (\Cref{figure:method:Overview}\textit{a}) using the estimated homography. 
Next, we forecast lost tracks in BEV space (\Cref{figure:method:Overview}\textit{b}) using our trajectory forecasting network (described in \Cref{sec:method:trajectoryForecasting}). Finally, we integrate forecasts into the online tracker (\Cref{sec:method:trajectoryIntegration}) while accounting for the uncertainty in estimated forecasts, and match new detections to existing tracks to resolve short- and long-term occlusions~(\Cref{figure:method:Overview}\textit{d}). 

\subsection{Data-driven Homography Estimation} 
\label{sec:method:homography}

\begin{figure}[ht]
    \centering
    \includegraphics[width=\textwidth]{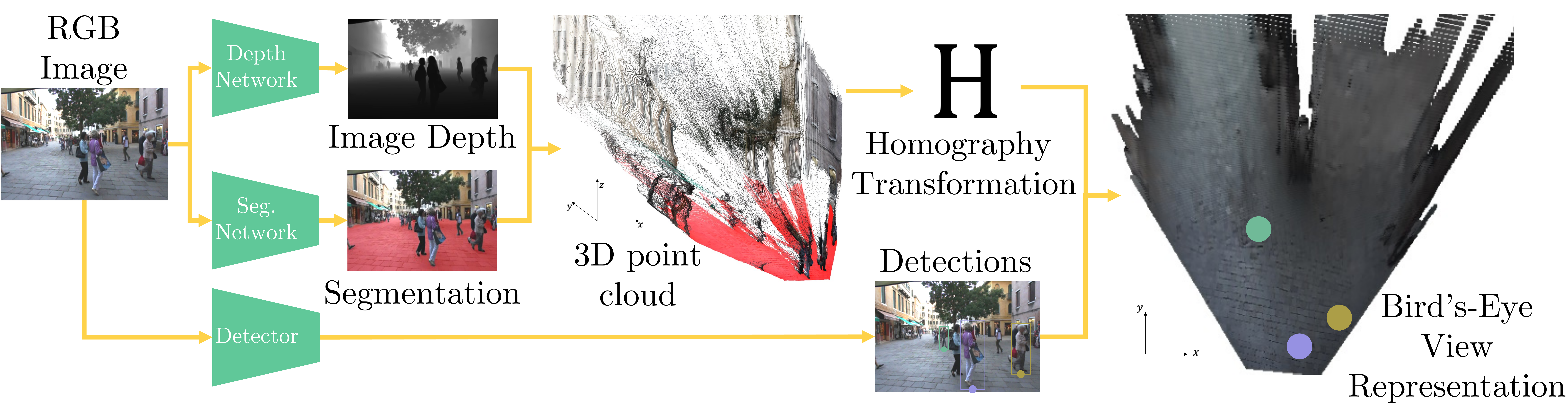}
    \caption{We estimate the homography $H$ for a sequence by reconstructing a 3D point cloud using a monocular depth estimator. We obtain ground image-to-point-cloud correspondences using a semantic segmentation model that masks ground pixels as needed to estimate the homography matrix. With the estimated homography matrix, we transform the bottom points of bounding boxes to 2D BEV coordinates.
    }
    \label{fig:method:homographyEstimation}
\end{figure}

For combining monocular object tracking and forecasting, we first need to transform object detections and tracks from the image sequence into points and trajectories in a bird's-eye view representation. 
Given a set of 2D object detections represented as bounding boxes localized in the image plane, we aim to find a homography $H$ that maps their bottom-center positions to their corresponding 2D BEV coordinates.
In~\Cref{fig:method:homographyEstimation}, we outline our homography estimation method. We first train a monocular depth estimator~\cite{bhat2020adabins} on a synthetic dataset~\cite{fabbri2021motsynth} to reconstruct a 3D point cloud (with estimated or known intrinsics) of the first frame of a static sequence. Then, we leverage the semantic segmentation network~\cite{wu2019detectron2} to mask, select, and fit a plane to the ground pixels. 
We estimate the normal vector of the ground plane in 3D and align the plane to the $XY$ plane. Then, we project ground points along the $z$-axis, leaving us with a pairwise correspondence between ground pixels in the image and a 2D position in BEV, as needed to estimate the homography between the two planes. We also linearize the homography transformation for pixel positions close to the plane's horizon to prevent the transformation from diverging (for more information, see~\Cref{sec:supp:linear}).

\PAR{Static camera.} We compute the homography only for the first frame of the sequence and use it throughout the sequence, making our pedestrian localization robust to temporal fluctuations of the depth estimator.  

\PAR{Moving camera.}
For moving camera sequences, we also need to account for the egomotion of the camera, which we estimate between consecutive frames as follows. 
First, we compute a frame-dependent homography $H_t$ for each frame. Then, we compute pairwise pixel-correspondences between (masked) ground pixels using optical flow~\cite{2021mmflow} and compute a translation vector between the two point sets (lifted to 3D via $H_t$). 

Empirically, we observe that estimating only translation (without rotation) yields more robust egomotion estimates. 
\subsection{Forecasting}
\label{sec:method:trajectoryForecasting}
Localization of object tracks in BEV enables us to leverage data-driven forecasting models beyond simple linear motion to reason for future trajectories. However, these models expect ground-truth fixed-size past trajectory observations, while our projected trajectories are noisy and of varying lengths. As discussed in \Cref{sec:forecasting_basics}, prediction models are optimized for metrics that incentivize a large number of predictions and minimize $L_2$ distance to ground-truth trajectories. It is thus unclear how different proposed concepts translate into real-world tracking scenarios. We, therefore, identify the main design patterns proposed in the forecasting community and verify their impact \textit{directly} in the context of \textit{forecasting to track}.

\PAR{Preprocessing.} 
Forecasting models encode trajectories using an LSTM encoder-decoder~\cite{Alahi16CVPR} architecture, which takes a fixed-size observed trajectory as input and predicts a future trajectory. We construct input trajectories from temporally consecutive detections of the same track ID localized in BEV. 
To account for localization noise, we smooth the noisy observations using the Kalman filter and linearly extrapolate trajectories into the past to get trajectories of the required fixed-size input length. 

\PAR{Trajectory forecasting design patterns.}
In our experimental setup, we build on the LSTM encoder-decoder architecture~\cite{Alahi16CVPR} and include the following key design patterns recently emerging in the forecasting community. 

\MPAR{Stochasticity.}
Stochastic trajectory predictors enable us to sample multiple plausible future trajectories to account for the uncertainty in future positions. 
We follow the approach by~\cite{Gupta18CVPR} and learn a generative GAN~\cite{nips04gan} model and train it with a best-of-many~\cite{best-of-many-sampling} loss. As a result, the network internally learns an observation-conditioned distribution of future trajectories, from which we can sample. 

\MPAR{Social Interactions.} 
Social interactions impact future motion: pedestrians adapt their trajectories on-the-fly to avoid collisions. Several methods~\cite{Gupta18CVPR,Alahi16CVPR,social_ways,bigat} account for interactions in the forecasting literature. These methods leverage pooling~\cite{Gupta18CVPR}, attention~\cite{sadeghian2018sophie}, or graph neural networks \cite{bigat} to provide social context (\ie, trajectories of surrounding agents) to the trajectory decoder. 
To answer whether modeling social interactions matters for tracking, we implement Social GAN (S-GAN)~\cite{Gupta18CVPR}, which uses pairwise interaction features between neighboring pedestrians by using max-pooling before they are passed to the decoder.

\MPAR{Multimodality and Diversity.}
While the aforementioned generative models learn a distribution over trajectories, they need to sample many trajectories to cover all modes present in the scene, as learning a multimodal posterior with a single GAN is difficult~\cite{GANmanifold}. To predict the scene's main modes with as few samples as possible, we implement a multi-generator GAN network~\cite{dendorfer21iccv}, extending the presented GAN architecture by training multiple decoder heads, where each decoder learns to focus on a particular model. As a result, we get a set of plausible but maximally separated predictions by sampling from these different generators.

\subsection{Tracking via Forecasting}
\label{sec:method:trajectoryIntegration}
We assume we have an online object tracker capable of tracking visible objects (\eg, bounding box regression-based tracker~\cite{Bergmann19ICCV}). As long as tracks are being updated with new detections, we consider them \textit{active} and keep them in the active set $\mathcal{S}_A$. Once we cannot associate a detection, a track becomes \textit{inactive} and is stored in the set of inactive tracks $\mathcal{S}_I$.  
For each frame $t$ the tracker outputs a set of tracks $\mathcal{O} = \left( o_1, \dots , o_M \right)$ with $o_i = \left(\mathtt{ID}_i,b_i, f_i\right)$ where $\mathtt{ID} \in \mathbb{N}^{+}$ represents the track identity, $b \in \mathbb{R}^4$ denotes a bounding box in pixel space (see \Cref{figure:method:Overview}a), and $f \in \mathbb{R}^D$ represents a $D$-dim feature vector encoding the appearance information obtained from a pre-trained convolutional network \cite{He16CVPR}. We localize bounding boxes in BEV coordinates $x \in \mathbb{R}^2$ using our estimated homography $H$.
\PAR{Quo Vadis?} 
If an object track becomes inactive (\ie, temporally lost), we move the active track into the memory bank and predict $k$ trajectories of length $\tau_{max}$ in BEV space using the trajectory forecasting model as described in \Cref{sec:method:trajectoryForecasting}. 
As long as the track is inactive and not yet matched, we move along the predicted trajectory and do not predict an entirely new trajectory in each frame. 

\PAR{Filtering and removing predictions.} 
\textit{No prediction can live forever.} When we use stochastic trajectory predictors with multiple predictions, we need to limit the number of inaccurate or obsolete predictions to decrease the chance of false re-association. 
In practice, we limit the lifetime of a prediction to a maximal lifetime of $\tau_{max}$ and try to filter out \textit{unlikely} forecasts.
We use spatial and social context to determine the \textit{freespace}~\cite{khurana2021detecting} in which objects should be visible to the camera. We assume that visible objects eventually are detected and, therefore, remove prediction branches that should be visible for more than $\tau_{vis}$ frames.

We consider an object as \textit{visible} if neither scene nor other pedestrians occlude the object. 
In particular, this means that the predicted BEV position lies in an area of the projected ground mask (shown in \Cref{figure:method:Overview}c) and has no bounding box overlap $\geq 0.25$ with any other object detection, closer to the camera. Relative order can be determined based on bottom bounding box coordinates for amodal detections. If all predictions from an inactive track are removed even before $\tau_{max}$, we also remove the entire track. 
\begin{figure}
    \centering
    \begin{subfigure}[t]{0.43\textwidth}
    \includegraphics[width=\textwidth]{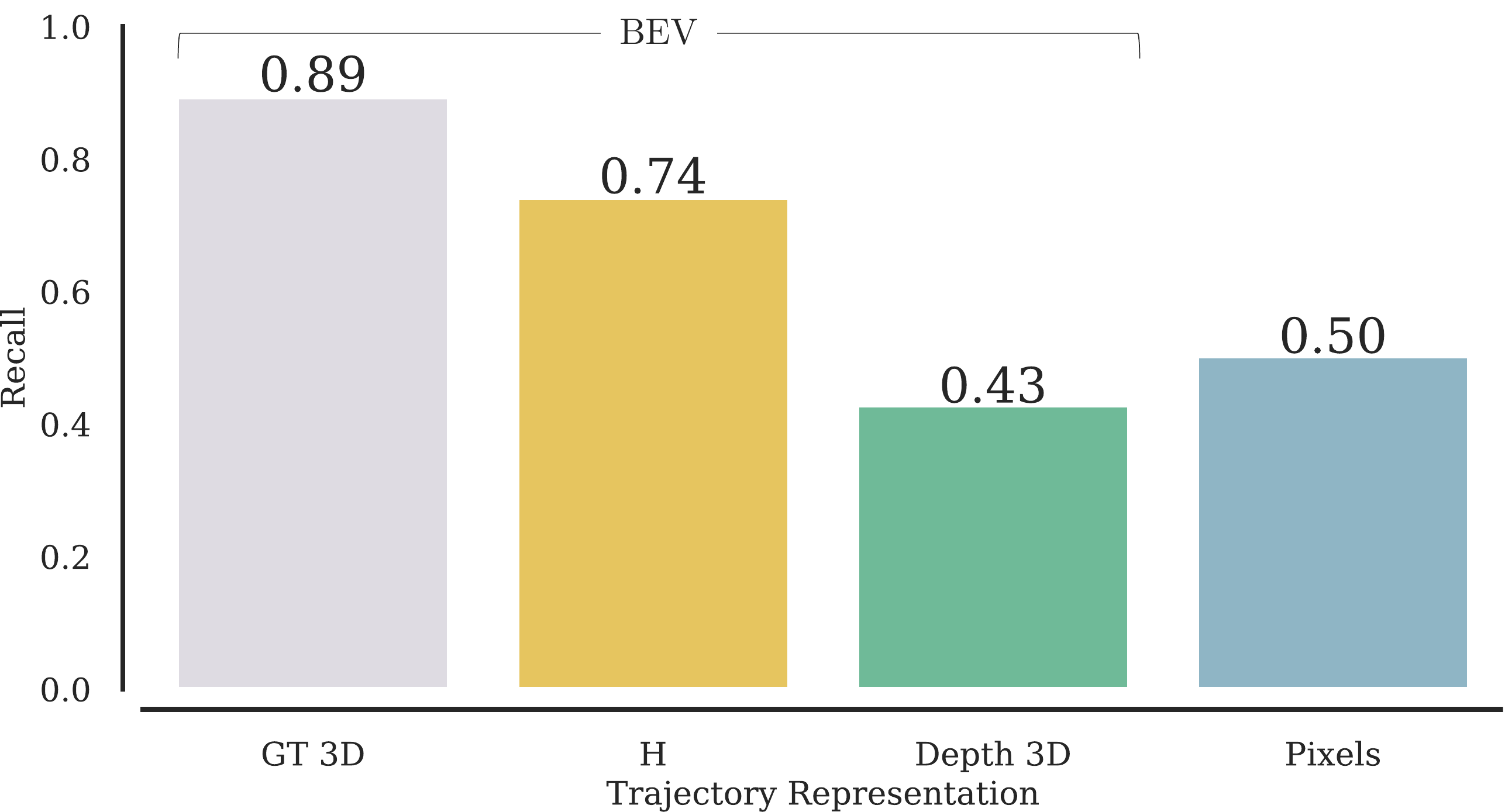}
    \caption{Overall recall (BEV and pixel-space).}
    \label{fig:experiment:3d_recall}
    \end{subfigure}
    \hfil
    \begin{subfigure}[t]{0.43\textwidth}
     \includegraphics[width=\textwidth]{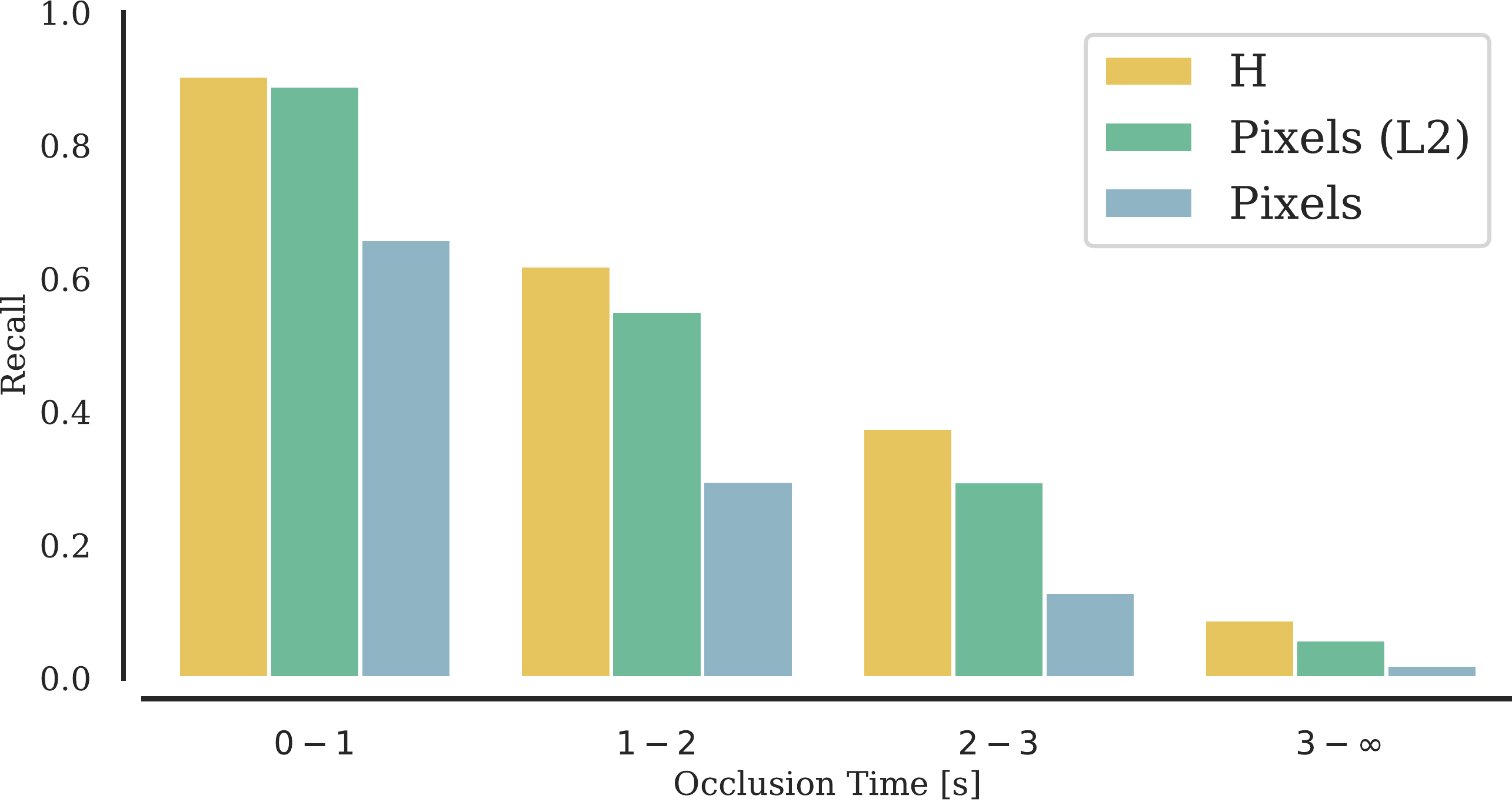}
    \caption{Recall \textit{wrt.} different occlusion lengths. }  \label{fig:experiments:3d_motion}
    \end{subfigure}
    \caption {Endpoint matching recall of predictions and GT trajectories using a linear motion predictor. A prediction is successfully matched when $\Delta_{\text{IoU}} > 0.5$ or $\Delta_{L_2}$ distance < $2m$. We also project the prediction back to the image for forecasts in the bird's-eye view. The model \textit{Pixel (L2)} predicts motion in pixel space and projects the endpoint into BEV for matching.  } 
    \label{fig:representation}
\end{figure}

\PAR{Matching predictions with new detections.} 
Given the trajectory forecasts, we match them with new detections via bi-partite matching, following the standard practice in tracking~\cite{Zhang2021IJCV, Wang20ECCV,Liu20ijcai,bewley16icip}. 
This boils down to computing association costs $c_{ij}$ between the predictions of an inactive track $i$ and new un-associated detections $j$:
\begin{equation}\label{eq:method:matching}
    c_{ij} = \left(\Delta_{\text{IoU}} + \text{max}\left(\tau_{L_2} - \Delta_{L_2}, 0 \right) \right) \cdot \left(\Delta_{\text{App}} \geq \tau_{\text{App}} \, \text{and}  \, \Delta_{\text{IoU}} \geq \tau_{\text{IoU}} \right), 
\end{equation}
where $\Delta_{\text{IoU}}$ is the IoU score between the two bounding boxes, $\Delta_{L_2}$ is the Euclidean distance between the prediction $i$ and a detection $j$ in BEV, and $\Delta_{\text{App}}$ represents the cosine distance between visual features $f_i$ and $f_j$. 
$\tau_{L_2}, \tau_{\text{IoU}}$, and $\tau_{\text{App}}$ denote thresholds for the matching. 
Therefore, we determine an association in BEV metric space and in the image domain using IoU bounding box overlap between the forecast and detected box. 

Matching tracks based on spatial distance in real space leads to high recall and reduces the number of ID switches (IDSW) but may also lead to an increase in ID transfer errors (IDTR), especially in crowded scenes with many new detections close to each other. While forecasting significantly narrows combinatorial search space for associations, verifying potential associations with an appearance model is still beneficial in practice. 
To decrease the number of wrong associations, we require a minimal visual similarity $\tau_{\text{App}}$ and minimum IoU overlap of the bounding boxes $\tau_{\text{IoU}}$ for close objects.
These thresholds serve as a filter of visually incompatible matches but do not add to the value of the cost function for the matching. 
In essence, we obtain a pre-selection of potential matching candidates by using the trajectory forecast and filter those if the appearance drastically deviates between the last observation and the new detection.

%
\section{Experimental Evaluation}
\label{sec:experimental}
In this section, we first discuss our evaluation test-bed, followed by an experimental study on bird's-eye-view (BEV) trajectory reconstruction (\Cref{sec:exp:birdEyeView}). Then, we analyze different forecasting design patterns applied to the domain of object tracking in BEV space and discuss the relevance of different model modules for tracking (\Cref{sec:exp:trajPred}). 
Afterward, we demonstrate how our approach can be used to improve several vision-based MOT methods on static sequences and to justify our design decisions. Finally, we show that our forecasting model can be used to establish new state-of-the-art on the real-world \MOTSEVENTEEN and \MOTTWENTY datasets (\Cref{sec:exp:benchmarkEval}). For visualization of our tracking method, we refer the reader to~\Cref{supp:sec:visualizations}. 

\PAR{Datasets.} We evaluate our trajectory prediction framework on different publicly available pedestrian tracking datasets, namely synthetic \MOTSYNTH~\cite{fabbri2021motsynth} and two real-world \MOTSEVENTEEN and \MOTTWENTY datasets~\cite{dendorfer20ijcv}. 
\textit{MOTSynth} is a large synthetic dataset for multi-object tracking. It provides $764$ diverse sequences with various viewpoints, lighting, and weather conditions. Importantly, it provides ground-truth depth information and 3D key points for pedestrians, allowing us to study the suitability of different methods for BEV trajectory reconstruction. \MOTSEVENTEEN~\cite{MOT17} and \MOTTWENTY~\cite{MOT20} are real-world tracking datasets commonly used to benchmark pedestrian tracking models. 
We use these datasets to evaluate our method on real-world recordings. 
For our experiments, we utilize the commonly used split of the \MOTSEVENTEEN training set, where the first half of each sequence is used for training and the second half for the evaluation~\cite{Zhuang20ECCV,byte_track,Wu21CVPR}. 

\PAR{Metrics.} 
For measuring the quality of the bird's-eye view reconstruction, we indirectly evaluate the quality of the reconstruction by evaluating the forecasting and tracking performance.

To compare different models for trajectory forecasting, we report the standard $L_2$ final displacement error (FDE) for top-k predictions for $2s$ and $4s$ prediction horizons (see \Cref{sec:forecasting_basics}). 

For multi-object tracking evaluation, we report higher-order tracking accuracy (HOTA)~\cite{luiten20ijcv}, with a focus on the association aspect of the task. To this end, we also report AssA, AssPr, and the number of ID switches IDSWs. 
 Additionally, we report  IDSW when the tracker loses an object and re-initiates a new track for the same object when it re-appears. We call these errors as  $\text{ID}^{\text{lost}} $. For metric discussion, we refer to \Cref{sec:problemdefinition}. Metrics labeled with either  $S$ (short) or $L$ (long) only consider prediction or occlusion lengths shorter or longer than $2s$, respectively

\PAR{Hyperparameters.} For all experiments, we use the following parameters for the matching of detections with inactive tracks: $\tau_{L_2} = 2.5m, \tau_{\text{App}} = 0.8,  \tau_{\text{IoU}} = 0.2$ The maximal lifetime of prediction is $\tau_{max} = 6s $ and maximal visibility  $\tau_{\text{vis}} = 1s$  before it is removed.
We refer the reader to \Cref{supp:sec:implementation} for further information on implementation details.

\PAR{Object trackers.}
We study and ablate our method on eight high-ranked state-of-the-art trackers of \MOTChallenge and refer to them as \textit{baseline}. We use BYTE~\cite{byte_track}, JDE~\cite{Wang20ECCV}, CSTrack~\cite{Zhuang20ECCV}, FairMOT~\cite{Zhang2021IJCV}, TraDes~\cite{Wu21CVPR}, QDTrack~\cite{Pang20CVPR}, CenterTrack~\cite{zhou2020tracking} and TransTrack~\cite{Sun20transtrack} for an evaluation on the \MOTSEVENTEEN validation set and BYTE and CenterTrack on the \MOTTWENTY training dataset. 
These trackers use ReID similarity and/or simple motion cues for bridging short-term occlusions. 
\subsection{Bird's-Eye View Estimation} 
\label{sec:exp:birdEyeView}
This section discusses different approaches to obtaining scene BEV representations of detected objects in the image for forecasting. 
\begin{table}[t]
     \caption{Which forecasting modules matter for tracking? Evaluated on \MOTSEVENTEEN validation set.}
    \input{tables/experiments_TrajectoryMOT}
    \label{tab:forecasting:MOT}
\end{table}
We work with static sequences of the \textit{MOTSynth} dataset (that provides depth maps used for evaluating and training a monocular depth estimator). With this, we test a linear motion model to gap occlusions of different durations, which we obtain by running a CenterTrack~\cite{zhou2020tracking} baseline tracker. We evaluate the ratio of successful matches between the target and the linear prediction and count a match to be successful if the IoU of the predicted bounding box is larger than $0.5$ or the $L_2$ distance in metric space is lower than $2m$. We get the predicted bounding box by translating the last observed bounding box by the predicted displacement in the image.

\PAR{Baselines.} We compare motion in (i) BEV and (ii) pixel space. We evaluate different approaches to localize trajectories: (a) using ground truth (GT) 3D coordinates orthographically projected to BEV (oracle), (b) the proposed homography estimation as described in \Cref{sec:method:homography}, and (c) directly using learned monocular depth estimates and resulting point clouds, followed by orthogonal projection of points these representing an object.

\PAR{Conclusions.}
As seen in \Cref{fig:representation}, GT 3D (oracle) based motion estimates solve $89.3 \%$ of the gaps, suggesting that the motion in the synthetic dataset is dominantly linear. Our proposed data-driven homography estimation approach only drops by $15\%$ compared to using ground-truth 3D keypoints. By contrast, estimating linear motion in pixels space only resolves $50.2 \%$, and using per-frame monocular depth estimates $43.1 \%$ of the occlusion gaps. This is likely because such depth estimates are not temporally stable. As can be seen in \Cref{fig:experiments:3d_motion}, this performance is especially apparent for longer occlusion gaps. 
Furthermore, we forecast motion in pixel space but transform the prediction into BEV for matching. While increasing performance compared to exclusive forecasting and matching in pixel space, we find that the results are inferior to predictions in BEV due to the distortion of projecting real motion into the image plane. We conclude that our proposed homography transformation is suitable for forecasting. 
\subsection{Trajectory Prediction Models}
\label{sec:exp:trajPred}
In this section, we evaluate different forecasting models and components (as discussed in \Cref{sec:method:trajectoryForecasting}). We compare a constant-velocity model (Kalman filter) in BEV and pixel space, an identity (static) model, and a stochastic GAN predictor generating $k=3$ and $k=20$ samples. Furthermore, we test predicting social interactions with GAN and the multimodal trajectories with MG-GAN in BEV space. 
\begin{tabularx}{\textwidth}{*{2}{>{\centering\arraybackslash}X}}
\vspace{2pt}
    \includegraphics[width=\linewidth,valign=m]{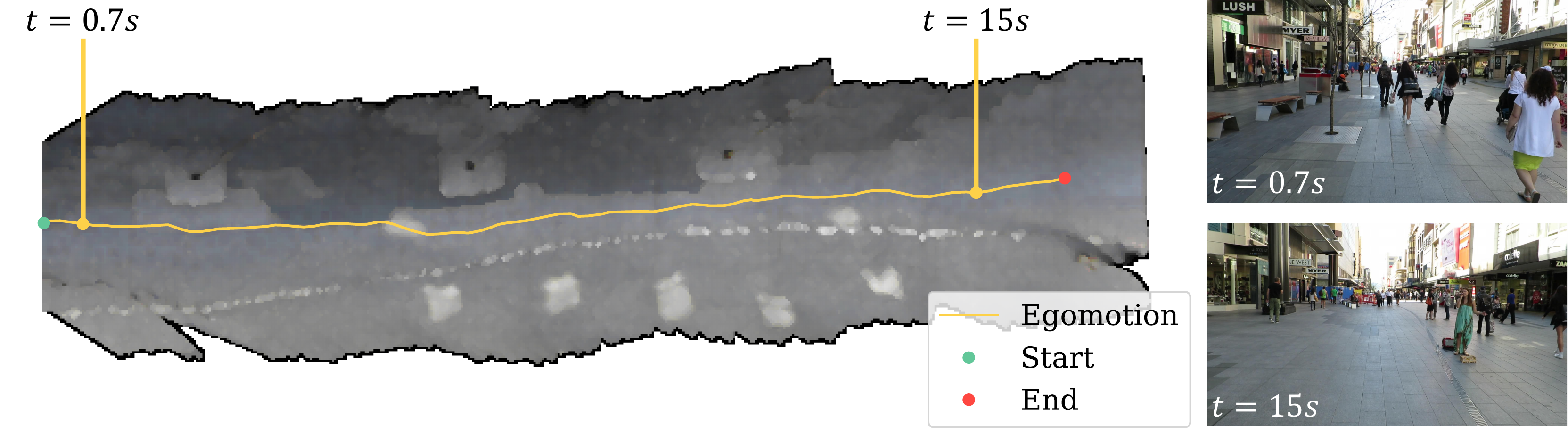}
&
\vspace{2pt}
    \parbox{\linewidth}{
    \input{tables/experiments_MOT_matching}
     } 
     \\
     \captionof{figure}{Visualization of BEV reconstruction for moving camera sequence and egomotion estimation.}
\label{fig:experiments:egomotion} 

&
     \captionof{table}{Ablation of matching prediction and effect of different thresholds $\tau$ on different tracking metrics.}
\label{tab:experiments:matching}
    \end{tabularx}
\PAR{Forecasting.} We observe in \Cref{tab:forecasting:MOT} that the linear model performs well for short-term windows ($0.69$ $\text{FDE}_S$), suggesting that linear motion is suitable for short occlusions. We also do not find a significant difference between GAN w/o social module (S-GAN). 
While FDE error suggests vanilla GAN ($k=20$) yields the best forecasting results ($0.65$ $\text{FDE}_S$ and $0.99$ $\text{FDE}_L$), but this configuration leads to the lowest association precision ($71.31$) and HOTA score ($53.81$) \wrt tracking performance.
This result suggests a misalignment of evaluation metrics used in forecasting and tracking; a better forecaster in terms of ADE/FDE does not necessarily lead to a better tracker. This is a known drawback of ADE/FDE metrics, which essentially measure only recall and not the precision of the forecasting output. Furthermore, this shows the careful trade-off between the number of predictions $k$ and the recall/precision of the predictions and tracking results.

\PAR{Tracking.} To investigate the effect on long-term occlusions, we focus on the change of $\text{ID}_L^{lost}$ for short ($t_{occl} \leq 2s$) and long ($t_{occl}> 2s$) occlusion gaps. As can be seen in \Cref{tab:forecasting:MOT}, even the static motion model solves $8.4 \%$ ($\text{ID}_L^{lost}$), as many occluded objects do not move. By modeling linear motion (Kalman filter in pixel space), we can improve short-term re-association for $0.59pp$ (long-term IDSW) over the static model. 
We focus the discussion on long occlusion gaps. In terms of the generative model, we observe that interaction-aware S-GAN ($ 8.64\%$ $\text{ID}_L^{lost}$) is on-par with vanilla GAN ($ 8.57\%$ $\text{ID}_L^{lost}$) for $k=3$; interestingly, both are below linear Kalman filter (BEV) performance, suggesting that these models suffer from low precision. Only MG-GAN, explicitly trained to generate multimodal trajectories, outperforms the linear model ($ 17.43\%$ $\text{ID}_L^{lost}$) and significantly outperforms vanilla GAN with only three samples. These conclusions generalize to tracking metrics.

\subsection{Tracking Evaluation}
\label{sec:exp:motionIntegration}
In this section, we study the impact of forecasting models on the valuation set's tracking performance. First, we discuss the impact of different design decisions on matching strategy, as explained in \Cref{sec:method:trajectoryIntegration}. 
\PAR{Trajectory matching.} First, we ablate the matching cost function (\Cref{eq:method:matching}). As can be seen in \Cref{tab:experiments:matching}, we find that a combination of $L_2$ and IoU without any threshold $\tau$ leads to the highest decrease in terms of $\text{ID}^{\text{lost}}$ ($-28.02 \% $) and overall highest association recall (AssRe) ($61.17$). However, this is at the cost of decreasing association precision (AssPr) ($-4.09$). We obtain the highest AssPr ($73.36$) by only relying on $L2$ matching and thresholding, however, at the loss of AssRe ($-0.74$). Adding appearance-based $\tau_{\text{App}}$ and IoU $\tau_{\text{IoU}}$ thresholds provide the best trade-off and overall highest AssA ($54.29$) and HOTA score ($54.27$) while still recovering $21 \%$ of lost trajectories. 
\PAR{Validation results.}
In \Cref{tab:tracking:MOT}, we present the performance of different state-of-the-art trackers on the \textit{static sequences} of \MOTSEVENTEEN-val and \MOTTWENTY-train (trained on \MOTSEVENTEEN), equipped with our trajectory forecasting model. As can be seen, our model brings stable improvements over all the key metrics: HOTA, AssA, and IDSW. Our trajectory prediction model consistently reduces IDSW for all models. This is also shown in \Cref{figure:teaser:occlusion} where we demonstrate that our forecasting model improves ID recall significantly for occlusion times $>1s$.
\begin{table}
    \caption{We improve tracking results of \textit{all} top-8 state-of-the-art models (MOT17 validation set and \MOTTWENTY training set). Differences to the baseline performance are shown in $\left( \cdot \right)$.}
    \input{tables/experiments_MOT}

    \label{tab:tracking:MOT}
\end{table}

While our focus was on sequences with stationary viewpoints, we show that our model is also applicable to sequences with moving cameras by estimating the camera's egomotion as described in~\Cref{sec:method:homography}. In \Cref{tab:supp:moving}, we present results on the \textit{moving sequences} of the MOT17 validation set (excluding sequence MOT17-05 for which the quality and consistency of our depth estimator was too low to construct time-consistent homographies). As can be seen, we improve $6$ out of $8$ trackers \wrt HOTA score and even improve CenterTrack \cite{zhou2020tracking} by 3.07 pp. 
We visualize the traversed BEV map of sequence MOT17-07 in \Cref{fig:experiments:egomotion}. 

\begin{table}
\centering
    \caption{Results of top-8 state-of-the-art models on dynamic scenes of the MOT17 validation set excluding MOT17-05. Differences in the baseline performance are shown in (·)}
    \input{tables/experiments_MOT_moving}

  \label{tab:supp:moving}
\end{table} 

\subsection{Benchmark Evaluation}
\label{sec:exp:benchmarkEval}

\begin{table}
\parbox{.48\linewidth}{
\centering
\caption{Comparison under the "private detector" protocol on \MOTSEVENTEEN test set.}
\input{tables/experiments_MOT_MOT17}

\label{tab:results:MOT17}
}
\hfill
\parbox{.48\linewidth}{
\centering
\caption{Comparison under the "private detector" protocol on \MOTTWENTY test set.}
\input{tables/experiments_MOT_MOT20}
\label{tab:results:MOT20}
}
\end{table}

In this section, we apply our method to state-of-the-art tracker ByteTrack~\cite{byte_track} and, by improving its long-term tracking capabilities, establish a new state-of-the-art on the \MOTSEVENTEEN \& \MOTTWENTY benchmarks. We evaluate our method in the \textit{private detection} regime, as these trackers use private detectors. 

In~\Cref{tab:results:MOT17}, we compare our \textit{QuoVadis} to the base tracker ByteTrack~\cite{byte_track} and compare both to \MOTSEVENTEEN benchmark published top-performers. We improve performance on key metrics overall top methods. Notably, we reduce the number of identity switches by 93 compared to \cite{byte_track} and establish a new state-of-the-art in terms of HOTA ($63.14$). We observe similar trends on \MOTTWENTY, where we improve over the base tracker ByteTrack~\cite{byte_track} by $+0.5$ in terms of IDF1 and reduce the number of identity switches by 36, similarly establishing a new state-of-the-art ($61.48$ HOTA).

\section{Remarks and Limitations}
\label{sec:remarks}
The paper primarily focuses on the conceptual work of building an entire pipeline from video to tracks studying different forecasting paradigms, and showing the benefit of leveraging trajectory forecasting in BEV for the tracking task. Nevertheless, we want to outline further remarks and limitations of our work.
\PAR{Model complexity.} Our model is complex, consisting of multiple sub-modules, as constructing object trajectories in BEV space based on a monocular video and trajectory forecasting are challenging problems.  
We foresee that future work will improve the end-to-end integration and efficiency of the algorithms. 
\PAR{Bird's-eye view reconstruction.}
A vital part of our approach is an accurate homography transformation that allows us to project the objects in the image into the 3D ground plane. However, the homography transformation depends on the quality of depth estimates, which makes the overall approach sensitive to errors in 3D localization. Future work will benefit from further development of time-consistent depth estimators. 
Furthermore, the presented trajectory prediction models do not yet account for the BEV localization uncertainty, which results from errors in the transformation or the simplified assumption of the ground plane. These limitations show the need to develop trajectory forecasting models that account for the localization uncertainties of the upstream tasks.

\section{Conclusion}
\label{sec:conclusion}
This paper presents a study on how to bridge the gap between real-world trajectory prediction and single-camera tracking. Throughout our paper, we identified challenges and solutions to leveraging real-world trajectory prediction to benefit single-camera tracking. In particular, we focus on resolving the re-identification of objects after long-term occlusions.
Here, we start from the first principles, questioning motion representation in pixel space and using a combination of models to construct a more accurate BEV representation of the scene. We find that the key component is a forecasting approach reasoning about multiple feasible future directions with a small set of multimodal forecasts. We can substantiate our conclusion by achieving new state-of-the-art performance on the \MOTSEVENTEEN and \MOTTWENTY datasets.

Ultimately, we have showcased a novel way of combining state-of-the-art trajectory prediction models and multi-object tracking task.
We have outlined a new way of thinking about motion prediction in tracking and motivating the beneficial symbiosis of both tasks. We hope that both fields start moving towards each other and incorporate the requirements and needs of each other. 

\PAR{Acknowledgments.} This research was partially funded by the Humboldt Foundation through the Sofja Kovalevskaja Award.

\newpage

{\small
\bibliographystyle{plain}
\bibliography{main}
}

\newpage

\appendix

\begin{center}
\bf{\huge Quo Vadis: Supplementary Material}
\\
\end{center}
The supplementary material complements our work with additional information on the bird's-eye-view reconstruction in \Cref{supp:sec:bevReconstruction}. Furthermore, we provide implementation and training details on different components and networks used in our method in \Cref{supp:sec:implementation}. Finally, we present visual examples as visualizations and videos in \Cref{supp:sec:visualizations}.
\input{supplementary}


\end{document}

%% file: macros.tex
\newcommand*{\eg}{\emph{e.g.}\@\xspace}
\newcommand*{\ie}{\emph{i.e.}\@\xspace}

\newcommand*{\wrt}{\emph{wrt.}\@\xspace}

\makeatletter
\newcommand*{\etc}{%
    \@ifnextchar{.}%
        {etc}%
        {etc.\@\xspace}%
}

\newcommand{\PAR}[1]{\vskip1pt \noindent {\bf #1~}}
\newcommand{\MPAR}[1]{\vskip1pt \noindent {\it #1~}}

\definecolor{codegreen}{rgb}{0.0,0.6,0.0}

\definecolor{scorec}{RGB}{31,119,180} 
\definecolor{objc}{RGB}{255,131,22} 
\definecolor{bgc}{RGB}{56,165,56} 
\definecolor{ensmc}{RGB}{214,39,40} 

\definecolor{darkyellow}{RGB}{253,209,73}
\definecolor{darkred}{RGB}{254, 71, 65} 
\definecolor{darkgrey}{RGB}{160, 160, 160}
\definecolor{darkgreen}{RGB}{98, 199, 153}

\newcommand{\MOTChallenge}{{\it MOTChallenge}\xspace}

\newcommand{\MOTSEVENTEEN}{{\it MOT17}\xspace}
\newcommand{\MOTTWENTY}{{\it MOT20}\xspace}
\newcommand{\MOTSYNTH}{{\it MOTSynth}\xspace}

%% file: tables/experiments_TrajectoryMOT.tex
\centering
\resizebox{\linewidth}{!}{
\begin{tabular}
{l|ccccc|rr|rrrrrr}
\toprule[1.5pt]

 \multirow{2}{*}{Model} &  \multirow{2}{*}{ \makecell{Nr. \\ Samples}} & \multirow{2}{*}{\makecell{Deter - \\ministic}} &
 \multirow{2}{*}{ \makecell{Stoch-\\ astic}} & \multirow{2}{*}{Social} & \multirow{2}{*}{ \makecell{Multi- \\ modal}} & 
 \multicolumn{2}{c}{Prediction} & 
 \multicolumn{5}{c}{Tracking}
 \\
 
 \cmidrule {7-14}
 &&&&&& FDE$_{S}$ $\downarrow$ & FDE$_{L}$ $\downarrow$ & HOTA  $\uparrow$& AssA $\uparrow$ & AssRe $\uparrow$ &  AssPr $\uparrow$ &  $\text{ID}^{\text{lost}}_S$ $\downarrow$& $\text{ID}^{\text{lost}}_L$  $\downarrow$     \\
\midrule[.5pt]
Baseline & & & & & & -- & -- &  50.71  & 46.87  & 51.80  & $\mathbf{78.11}$   & 0 \% & 0 \% \\ 
Static  &1 & \checkmark && && 1.59 &2.09 &  53.84  & 53.51  & 60.04  & 72.95  & -14.77 \% & -8.40 \% \\ 
Kalman Filter (pixel) &1 & \checkmark   &&  & & --    & --     & 54.08  & 54.02  & 60.45  & 72.81  & -$\mathbf{22.37} $ \% & -8.99 \% \\ 

Kalman Filter  & 1 & \checkmark & & & 
& 0.69 & 1.23  &54.11  & 54.04  & 60.75  & 71.73  & -19.50 \% & -16.07 \% \\ 
GAN  & 3&  &  \checkmark  & & 
&  0.85     & 1.26  &
54.43  & 54.61  & 61.11  & 73.21  & -17.99 \% & -8.64 \% \\ 
GAN & 20  & &  \checkmark && & $\mathbf{0.65} $  & $\mathbf{0.99}$  & 53.81  & 53.40  & 60.45  & 71.31  & -18.03 \% & -15.63 \% \\ 
S-GAN & 3  &  & \checkmark& \checkmark && 0.87  & 1.21   &    $\mathbf{54.52}$  & 54.78  & 61.22  & 73.28  & -16.92 \% & -8.57 \% \\ 
MG-GAN & 3   &  & \checkmark &  & \checkmark& 0.67&  1.03 & $\mathbf{54.52}$  & $\mathbf{ 54.80 } $ &$\mathbf{ 61.35 }$ & 73.13  & -21.19 \% & -$\mathbf{17.43} $ \% \\ 
\bottomrule[1.5pt]
\end{tabular}}

%% file: tables/experiments_MOT_matching.tex
\resizebox{\linewidth}{!}{
\begin{tabular}{C{0.65cm}C{0.5cm}|C{0.5cm}C{0.5cm}|R{0.9cm}R{0.9cm}R{0.9cm}R{0.9cm}r}
  \multicolumn{2}{c}{Scores} &  \multicolumn{2}{c}{Threshold} & 
 \multirow{2}{*}{HOTA $\uparrow$ } &  \multirow{2}{*}{AssA$\uparrow$  } &  \multirow{2}{*}{ AssRe $\uparrow$ } &  \multirow{2}{*}{ AssPr $\uparrow$ } &  \multirow{2}{*}{ $\text{ID}^{\text{lost} } \downarrow $  } \\
 $L_2$ & IoU  & $\tau_{\text{IoU}}$ & $\tau_{\text{App}}$  & & & & \\
\midrule
 \checkmark& & &      & 53.89   & 53.56   & 60.43   & 72.21   & -16.18 \%  \\ 
\checkmark &  &   & \checkmark   & 53.89   & 53.57   & 60.51   & 71.69   & -16.26  \% \\ 
\checkmark &  & \checkmark  & \checkmark  & 54.10   & 53.92   & 60.43   & $\mathbf{73.36}$   & -16.84 \%  \\ 
 & \checkmark &&    & 54.13   & 54.01   & 60.97   & 72.00   & -24.06   \% \\ 
    \checkmark & \checkmark & & &  53.75   & 53.35   & $\mathbf{61.17} $   & 69.27   & -$\mathbf{28.02  \%}$ \\ 

 \checkmark & \checkmark & $\checkmark$  &  & 53.97   & 53.75   & 61.08   & 70.73   & -26.93 \%   \\ 
 \checkmark & \checkmark & &\checkmark &   54.06   & 53.92   & 61.07   & 71.01   & -21.40 \%  \\ 
 \checkmark & \checkmark & $\checkmark$  & $\checkmark$  & $\mathbf{54.27}$   & $\mathbf{54.29} $    & 61.08   & 72.36   & $-20.53  \%$   \\ 
\end{tabular}
}

%% file: tables/experiments_MOT.tex
\centering
\resizebox{\linewidth}{!}{
\begin{tabular}{llllllllllll}
\toprule[1.5pt]
& \multicolumn{8}{c}{MOT17 (val, static scenes)} &  \multicolumn{2}{c}{MOT20 (train)} \\
\cmidrule(lr){2-9}
\cmidrule(lr){10-11}
&  BYTE~\cite{byte_track}   &  CenterTrack~\cite{zhou2020tracking}   &  CSTrack~\cite{Zhuang20ECCV}   &  FairMOT~\cite{Zhang2021IJCV}   &  JDE~\cite{Wang20ECCV}   &  TraDeS~\cite{Wu21CVPR}   &  TransTrack~\cite{Sun20transtrack}& QDTrack~\cite{Pang20CVPR}   &  BYTE~\cite{byte_track}   &  CenterTrack~\cite{zhou2020tracking}    \\
HOTA & 71.36  \textcolor{codegreen}{ (+0.21)}& 61.78  \textcolor{codegreen}{ (+3.56)}& 61.60  \textcolor{codegreen}{ (+0.43)}& 58.42  \textcolor{codegreen}{ (+0.09)}& 51.06  \textcolor{codegreen}{ (+0.20)}& 62.45  \textcolor{codegreen}{ (+0.67)}& 60.68   (-0.23)&58.87  \textcolor{codegreen}{ (+0.54)} &  56.85  \textcolor{codegreen}{ (+0.06)}& 32.71  \textcolor{codegreen}{ (+0.62)} \\ 
AssA & 73.96  \textcolor{codegreen}{ (+0.49)}& 66.18  \textcolor{codegreen}{ (+7.54)}& 63.84  \textcolor{codegreen}{ (+0.8)}& 59.21  \textcolor{codegreen}{ (+0.37)}& 54.36  \textcolor{codegreen}{ (+0.45)}& 67.41  \textcolor{codegreen}{ (+1.6)}& 63.47   (-0.49)&  60.14  \textcolor{codegreen}{ (+1.22)}& 53.97  \textcolor{codegreen}{ (+0.20)}& 28.94  \textcolor{codegreen}{ (+1.34)} \\ 

AssRe & 79.21  \textcolor{codegreen}{ (+0.66)}& 69.66  \textcolor{codegreen}{ (+8.38)}& 69.15  \textcolor{codegreen}{ (+1.07)}& 64.31  \textcolor{codegreen}{ (+0.5)}& 60.82  \textcolor{codegreen}{ (+0.89)}& 73.1  \textcolor{codegreen}{ (+2.43)}& 69.19  \textcolor{codegreen}{ (+0.02)}& 65.31  \textcolor{codegreen}{ (+2.06)}& 59.89  \textcolor{codegreen}{ (+0.4)}& 34.34  \textcolor{codegreen}{ (+5.12)} \\ 

AssPr & 83.11   (-0.67)& 81.75   (-5.47)& 77.79   (-2.28)& 74.45   (-1.71)& 68.9   (-2.35)& 80.0   (-1.91)& 79.53   (-1.68)& 77.4   (-2.98)& 68.65   (-5.24)& 52.37   (-21.06) \\ 
IDSW & 84  \textcolor{codegreen}{ (-3)}& 137  \textcolor{codegreen}{ (-146)}& 269  \textcolor{codegreen}{ (-28)}& 198  \textcolor{codegreen}{ (-12)}& 316  \textcolor{codegreen}{ (-19)}& 106  \textcolor{codegreen}{ (-32)}& 112  \textcolor{codegreen}{ (-3)}&219  \textcolor{codegreen}{ (-34)}& 1815  \textcolor{codegreen}{ (-78)}& 5240  \textcolor{codegreen}{ (-2700)} \\ 
MOTA & 80.09  \textcolor{codegreen}{ (+0.01)}& 70.77  \textcolor{codegreen}{ (+0.39)}& 71.31  \textcolor{codegreen}{ (+0.05)}& 71.82  \textcolor{codegreen}{ (+0.05)}& 59.57  \textcolor{codegreen}{ (+0.06)}& 70.93  \textcolor{codegreen}{ (+0.09)}& 69.5  \textcolor{codegreen}{ (+0.01)}& 69.61  \textcolor{codegreen}{ (+0.08)}& 73.38  \textcolor{codegreen}{ (+0.0)}& 47.57  \textcolor{codegreen}{ (+0.24)} \\ 

IDF1 & 82.92  \textcolor{codegreen}{ (+0.42)}& 74.46  \textcolor{codegreen}{ (+7.13)}& 74.16  \textcolor{codegreen}{ (+0.95)}& 73.93  \textcolor{codegreen}{ (+0.59)}& 65.01  \textcolor{codegreen}{ (+1.27)}& 76.36  \textcolor{codegreen}{ (+1.21)}& 71.46  \textcolor{codegreen}{ (+0.02)}& 70.41  \textcolor{codegreen}{ (+0.77)}& 72.47  \textcolor{codegreen}{ (+0.37)}& 45.85  \textcolor{codegreen}{ (+4.13)} \\ 

IDR & 78.61  \textcolor{codegreen}{ (+0.39)}& 65.25  \textcolor{codegreen}{ (+6.25)}& 67.53  \textcolor{codegreen}{ (+0.87)}& 66.23  \textcolor{codegreen}{ (+0.53)}& 56.08  \textcolor{codegreen}{ (+1.09)}& 67.12  \textcolor{codegreen}{ (+1.06)}& 61.39  \textcolor{codegreen}{ (+0.01)}& 62.17  \textcolor{codegreen}{ (+0.68)}& 66.44  \textcolor{codegreen}{ (+0.34)}& 35.87  \textcolor{codegreen}{ (+3.23)} \\ 

IDP & 87.72  \textcolor{codegreen}{ (+0.44)}& 86.71  \textcolor{codegreen}{ (+8.3)}& 82.23  \textcolor{codegreen}{ (+1.05)}& 83.65  \textcolor{codegreen}{ (+0.67)}& 77.31  \textcolor{codegreen}{ (+1.51)}& 88.55  \textcolor{codegreen}{ (+1.4)}& 85.47  \textcolor{codegreen}{ (+0.02)}& 81.17  \textcolor{codegreen}{ (+0.89)}& 79.7  \textcolor{codegreen}{ (+0.41)}& 63.53  \textcolor{codegreen}{ (+5.72)} \\ 
\bottomrule[1.5pt]
\end{tabular}}

%% file: tables/experiments_MOT_moving.tex
\centering
\resizebox{\linewidth}{!}{
\begin{tabular}{lllllllll}
\toprule[1.5pt]
& \multicolumn{8}{c}{MOT17 (val, moving scenes)} \\
\cmidrule(lr){2-9}

&  BYTE~\cite{byte_track}   &  CenterTrack~\cite{zhou2020tracking}   &  CSTrack~\cite{Zhuang20ECCV}   &  FairMOT~\cite{Zhang2021IJCV}   &  JDE~\cite{Wang20ECCV}   &  TraDeS~\cite{Wu21CVPR}   &  TransTrack~\cite{Sun20transtrack}  & 
QDTrack~\cite{Pang20CVPR} \\ 
HOTA & 60.08  \textcolor{codegreen}{ (+0.02)}& 51.77  \textcolor{codegreen}{ (+3.07)}& 54.51   (0.0)& 56.1   (0.0)& 52.14  \textcolor{codegreen}{ (+1.47)}& 53.36  \textcolor{codegreen}{ (+1.27)}& 52.7  \textcolor{codegreen}{ (+0.28)}& 52.28  \textcolor{codegreen}{ (+0.76)} \\ 

AssA & 60.44  \textcolor{codegreen}{ (+0.03)}& 53.18  \textcolor{codegreen}{ (+6.49)}& 59.04  \textcolor{codegreen}{ (+0.0)}& 61.15   (0.0)& 55.32  \textcolor{codegreen}{ (+3.06)}& 54.08  \textcolor{codegreen}{ (+2.44)}& 51.99  \textcolor{codegreen}{ (+0.54)}& 53.71  \textcolor{codegreen}{ (+1.6)} \\ 

AssRe & 66.53   (-0.0)& 58.21  \textcolor{codegreen}{ (+8.32)}& 63.35  \textcolor{codegreen}{ (+0.0)}& 65.87   (0.0)& 61.52  \textcolor{codegreen}{ (+3.75)}& 59.86  \textcolor{codegreen}{ (+3.1)}& 59.13  \textcolor{codegreen}{ (+0.79)}& 61.8  \textcolor{codegreen}{ (+3.82)} \\ 

AssPr & 78.29  \textcolor{codegreen}{ (+0.09)}& 76.98   (-4.66)& 80.46   (-0.0)& 79.21   (0.0)& 73.5   (-0.98)& 75.52   (-3.08)& 72.45   (-0.32)& 72.53   (-4.76) \\ 

IDSW & 54   (+1)& 131  \textcolor{codegreen}{ (-62)}& 97  \textcolor{codegreen}{ (-5)}& 86   (0)& 122  \textcolor{codegreen}{ (-10)}& 99  \textcolor{codegreen}{ (-11)}& 120  \textcolor{codegreen}{ (-5)}& 71  \textcolor{codegreen}{ (-7)} \\ 

MOTA & 72.54   (-0.01)& 59.46  \textcolor{codegreen}{ (+0.46)}& 60.68  \textcolor{codegreen}{ (+0.04)}& 63.78   (0.0)& 60.52  \textcolor{codegreen}{ (+0.07)}& 64.13  \textcolor{codegreen}{ (+0.08)}& 63.64  \textcolor{codegreen}{ (+0.04)}& 60.21  \textcolor{codegreen}{ (+0.05)} \\ 

IDF1 & 73.11   (0.0)& 63.48  \textcolor{codegreen}{ (+5.76)}& 70.69   (0.0)& 73.1   (0.0)& 68.23  \textcolor{codegreen}{ (+1.85)}& 67.72  \textcolor{codegreen}{ (+2.29)}& 64.08  \textcolor{codegreen}{ (+0.79)}& 65.81  \textcolor{codegreen}{ (+2.86)} \\ 
 
\bottomrule[1.5pt]
\end{tabular}}

%% file: tables/experiments_MOT_MOT17.tex
\centering
\resizebox{\linewidth}{!}{
\begin{tabular}{l|lllll}

\toprule[1.5pt]
Tracker & HOTA & IDF1 & MOTA & IDSW & AssA \\
\midrule
\text{ReMOT \cite{re_mot}}                 
    & 59.73 & 71.99 & 77.01 & 2853 & 57.08  \\
\text{CrowdTrack \cite{9663836}}
    & 60.26 & 73.62 & 75.61 & 2544 & 59.26  \\
\text{TLR \cite{wang:cvpr:21}}                      
    & 60.72 & 73.58 & 76.48 & 3369 & 58.88  \\
\text{MAA \cite{Stadler_2022_WACV}}         
    & 61.98 & 75.88 & 79.36 & 1452 & 60.16  \\
\text{ByteTrack \cite{byte_track}}         
    & 63.05 & 77.30 & 80.25 & 2196 & 61.97  \\
\bottomrule[1.5pt]

QuoVadis (Ours)
    & $\mathbf{63.14}$ & $\mathbf{77.71}$ & $\mathbf{80.27}$ & $\mathbf{2103}$ & $\mathbf{62.07}$

\end{tabular}
}

%% file: tables/experiments_MOT_MOT20.tex




\centering
\resizebox{\linewidth}{!}{
\begin{tabular}{l|lllll}

\toprule[1.5pt]
Tracker     & HOTA & IDF1 & MOTA & IDSW & AssA \\
\midrule

\text{FairMOT \cite{Zhang2021IJCV}}
    & 54.42 & 68.44 & 59.57 & 1881 & 56.6  \\
\text{CrowdTrack \cite{9663836}}
    & 54.95 & 68.24 & 70.68 & 3198 & 52.57  \\    
\text{MAA \cite{Stadler_2022_WACV}}        
    & 57.28 & 71.15 & 73.90 & 1331 & 55.14  \\
\text{ReMOT \cite{re_mot}}                 
    & 61.15 & 73.14 & 77.42 & 1789 & 58.68  \\
\text{ByteTrack \cite{byte_track}}         
    & 61.34 & 75.20 & 77.76 & 1223 & 59.55  \\
\bottomrule[1.5pt]

QuoVadis (Ours)
    & $\mathbf{61.48}$ & $\mathbf{75.70}$ & $\mathbf{77.77}$ & $\mathbf{1187}$ & $\mathbf{59.87}$

\end{tabular}
}

%% file: supplementary.tex
\section{Information on Bird's-Eye View Reconstruction}
\label{supp:sec:bevReconstruction}
The paper presents our approach to constructing a bird's-eye-view (BEV) representation for a static tracking sequence. Here, we extend the explanation by adding a description of moving cameras and how we linearize the homography transformation for farther objects to avoid enormous distances and unrealistic velocities.

\subsection{Linearization of  Homography}
\label{sec:supp:linear}
\begin{figure}
    \centering
    \begin{subfigure}[b]{0.58\textwidth}
    \includegraphics[width=\textwidth]{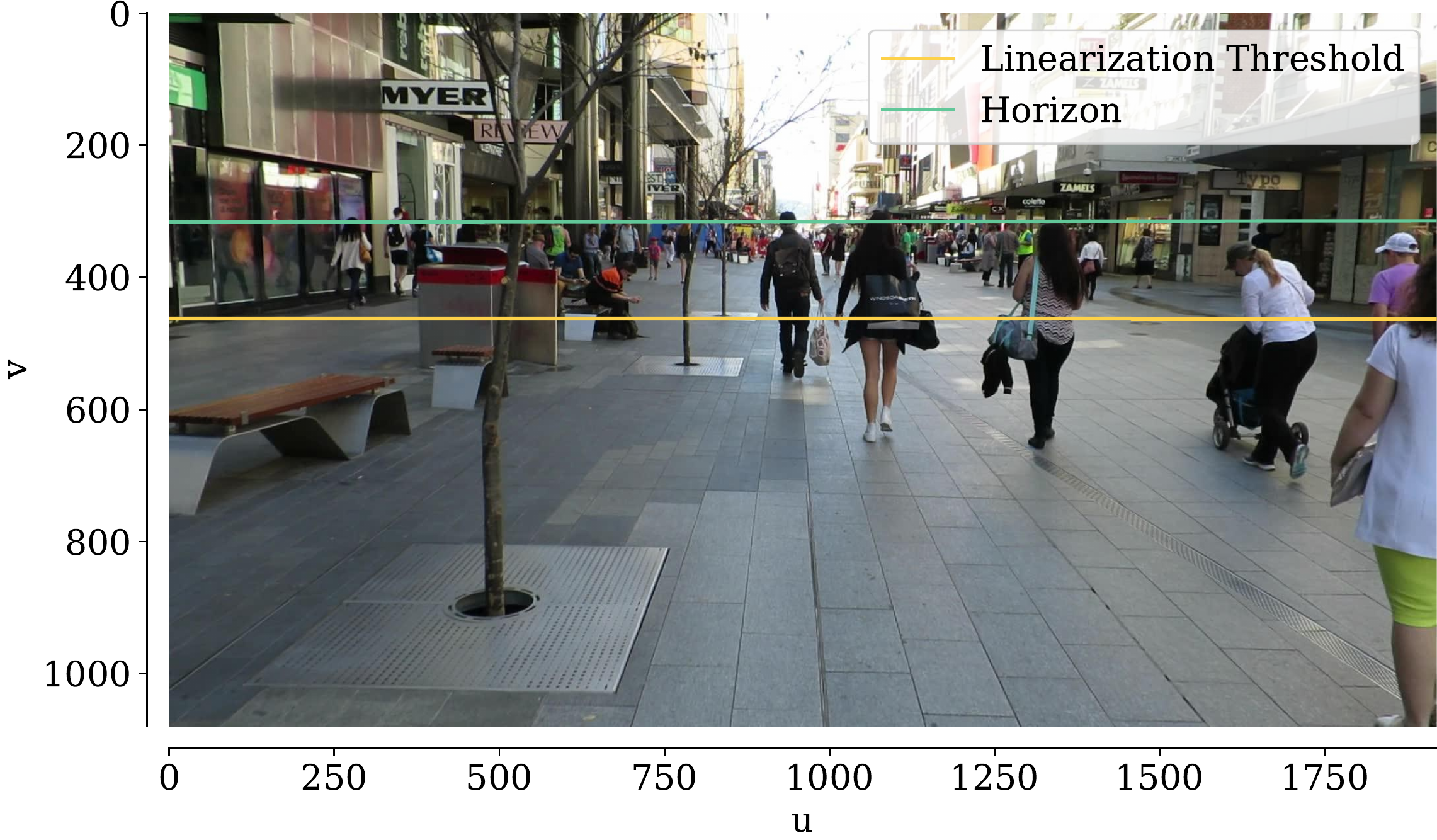}
    \caption{Horizon and Linear Threshold in Scene.}
    
    \label{fig:supp:scene}
    \end{subfigure}
    \hfil
    \begin{subfigure}[b]{0.41\textwidth}
     \includegraphics[width=\textwidth]{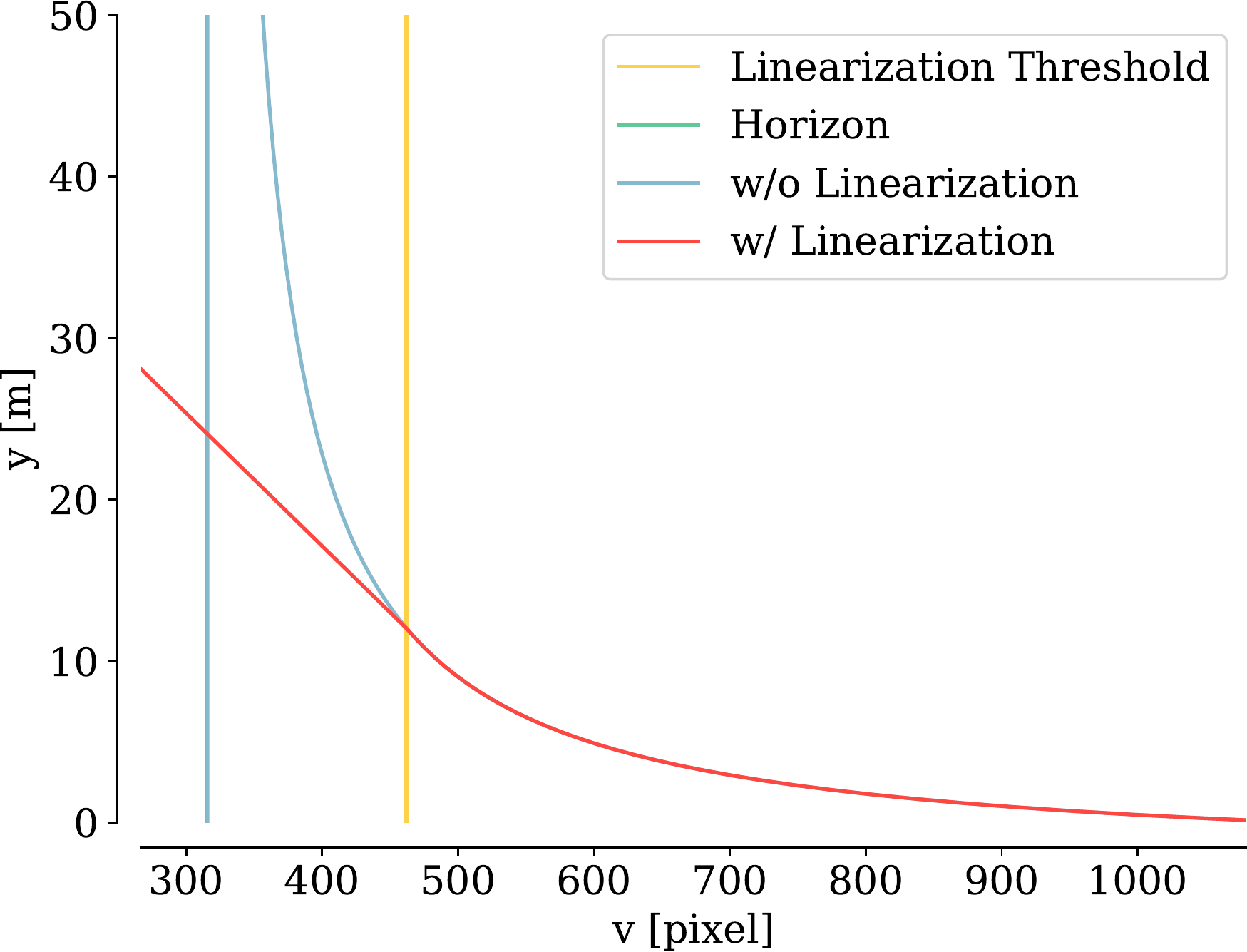}
    \caption{Linearization of BEV coordinates.}  \label{fig:supp:linearization}
    \end{subfigure}
    \caption{Demonstration of horizon and linearization threshold for sequence image. Linearization of homography transformation is necessary to prevent enormous distances in the transformed coordinates and unrealistic velocities.} 
    \label{fig:supp:representation}
\end{figure}

To get a bird's-eye-view (BEV) representation of the tracking scene, we estimate the homography  $H$ between the image and the ground plane. Hence, the homogenous pixel positions transform accordingly to 
\Cref{eq:homography:transformation} as follows:
\begin{equation} \label{eq:homography:transformation}
    s \cdot \begin{pmatrix} x  \\ y \\ 1 \end{pmatrix} = H \cdot \begin{pmatrix} p_x \\ p_y \\ 1 \end{pmatrix}.
\end{equation}
This approach assumes that objects move on a perfect plane and object's position in the image is represented as the bottom mid-point of the object's bounding box. Depending on the perspective transformation of the camera, we find that minor changes in pixel value lead to enormous distances in BEV. 
Given a homography matrix: 
\begin{equation} 
H = \begin{pmatrix}
h_{11} & h_{12} & h_{13} \\
h_{21} & h_{22} & h_{23} \\
h_{31} & h_{32} & h_{33} 
\end{pmatrix},
\end{equation}
the BEV coordinate $y$ is computed as:

\begin{equation} \label{eq:supp:homographyY}
y = \frac{h_{21} \cdot p_x + h_{22} \cdot p_y  + h_{23}}{h_{31} \cdot p_x + h_{32} \cdot p_y +  h_{33}}. 
\end{equation}
As the denominator in~\Cref{eq:supp:homographyY} is approaching zero, the $y$-coordinate grows hyperbolically. 
This behavior is undesired for trajectory prediction because these large jumps in the object's position result in unrealistic velocities for the object. Therefore, we define a threshold for which we linearly extrapolate the transformation such that the transformed distance between two neighboring pixel points is maximal $0.2m$ as shown as a red line in~\Cref{fig:supp:linearization}. This formulates the condition as 
\begin{equation} \label{eq:supp:threshold}
    \lVert \frac{h_{21} \cdot p_x + h_{22} \cdot p_y  + h_{23}}{h_{31} \cdot p_x + h_{32} \cdot p_y +  h_{33}} - 
    \frac{h_{21} \cdot p_x + h_{22} \cdot ( p_y + 1)  + h_{23}}{h_{31} \cdot p_x + h_{32} \cdot ( p_y + 1) +  h_{33}}   \rVert \leq 0.2m 
\end{equation}
We call the $p_y$ value for which the inequality~\Cref{eq:supp:threshold} is equal,  the linearization threshold $p_y^T$. 
The pixel point where the denominator of~\Cref{eq:supp:homographyY} becomes $0$ is called the horizon because no point on the plane is projected on a lower point in the image. 

To prevent this hyperbolic growth for image points closer to the horizon, we linerarize~\Cref{eq:supp:homographyY} around $ p_y^T$ and apply the linear transformation for all $ p_y \leq p_y^T$ as shown in~\Cref{fig:supp:linearization}. Thus,  we stabilize the distance between two points to prevent very unrealistic velocities, which would make the transformed values pointless. To transform from pixel space to BEV and back, we also inverse the linearized transformation to get a one-to-one mapping.

\section{Implementation Details}
\label{supp:sec:implementation}
In this section, we provide additional information on the implementation of our method and its key components. The source code is available at \url{https://github.com/dendorferpatrick/QuoVadis}.

\subsection{Synthetic Training Data}
For training the trajectory predictor (\Cref{sec:supp:trajectoryPredictor}) and the depth estimator (\Cref{sec:supp:depthEstimator}) we use the MOTSynth dataset \cite{fabbri2021motsynth} which provides ground truth 3D positions of objects and image depth information. We use the split suggested by Fabbri \textit{et al.}~\cite{fabbri2021motsynth} with 576 sequences in the training set and 192 in the validation set.
\subsection{Trajectory Predictor}
\label{sec:supp:trajectoryPredictor}
For our trajectory model, we use the implementation of MG-GAN~\cite{dendorfer21iccv}. For studying the effect of modeling social interactions on tracking, we implement a social max-pooling module following S-GAN~\cite{Gupta18CVPR}.

\PAR{Model.} The trajectory prediction model generates a set of $K$ future trajectories $\{\hat{Y}^k_i\}_{k=1, \dots, K}$ with $t \in \left[t_{obs}+ 1, t_{pred} \right]$ given the input trajectory $X_i$ with $t \in \left[t_{1}, t_{obs} \right]$ for each pedestrian $i$. We use $t_{obs} = 8$ observation steps and $t_{pred} = 12$ prediction steps as default for training the model. However, input and output length can vary depending on the observed tracks during inference for the tracking model.

For the multimodal MG-GAN implementation, we use $n_G = 3$ generators from which we sample one prediction from each generator during inference.

\PAR{Training.} We construct trajectories of the MOTSynth data with $8$ observation and $12$ prediction steps, each step being $0.4s$. 
The entire model is trained in a GAN framework using a prediction model and a discriminator network. 
We train the network on the entire train dataset over $200$ epochs, with a learning rate $\lambda  = 10^{-3}$,  and using a batch size scheduler~\cite{smith18batchsize}.

\subsection{Depth Estimator}
\label{sec:supp:depthEstimator}
Depth estimation is a crucial part of the BEV estimation in our model. 
Therefore, we use a vision transformer-based \cite{dosovitskiy2021image} network \cite{bhat2020adabins} for monocular dense depth estimation.

\PAR{Model. } The transformer-based model regresses the depth prediction as a linear combination of depth range bins of adaptive size. The network encoder-decoder extracts visual features from the image, which are passed to the mVit block. mVit is a lightweight vision transformer based on \cite{dosovitskiy2021image}. The model applies an MLP on top of the mVit's output, predicting the size of the bins for the depth range. The encoder computes the weights of the bins by passing the features through multiple convolutional layers with a final softmax non-linear activation function.

\PAR{Training.}  The network trains on the synthetic MOTsynth dataset to leverage a large number of tracking scenes of varying perspectives, weather, and light conditions. To better generalize to real-world data, we augment the scenes with ground reflections by mirroring surfaces in the image. This results in better performance, especially for the indoor MOT sequences, with ground reflections. We find the model trained on synthetic data performs well on real data even without fine-tuning. 

To increase the default model depth map resolution from $640 \times 480$ to $960 \times 576$ we grow the transformer positional embedding vector size from 500 to 1000. 

 We trained the model using AdamW optimizer \cite{loshchilov2019decoupled} with weight decay $10^{-2}$. Following \cite{smith2018superconvergence}, the maximum learning rate $\lambda_{\text{max}}$ was set to $3.5 \times 10^{-4}$ with linear warm-up from $\frac{1}{4} \lambda_{\text{max}} \to \lambda_{\text{max}}$ for the first 30\% of the iterations followed by cosine annealing to $\frac{3}{4}\lambda_{\text{max}}$. We trained the model for 20 epochs with an image resolution of $960 \times 576$ on a training split of MOTSynth dataset with the batch size of 8 on 4 $RTX8000$ for one week. Then, we trained the model for 30 epochs with an image resolution of $960 \times 576$ on the full MOTSynth dataset with a batch size of 8 on 4 $RTX8000$ for ten days. 

\subsection{Image Segmentation}
\label{sec:supp:segmentation}
We run a pre-trained Detectron2~\cite{wu2019detectron2} segmentation network to get the segmentation masks of the tracking scene images. Explicitly, we use the pre-trained COCO Panoptic Segmentation model with Panoptic FPN~\cite{wu2019detectron2}. The model outputs semantic labels for $134$ COCO classes and panoptic object ids, which are irrelevant to our task.

In our model, we use segmentation labels to mask ground pixels of the scene. Therefore, we combine the following COCO classes to our ground class:  $\mathtt{pavement}, \mathtt{road},\mathtt{platform}, \mathtt{floor}, \mathtt{floor-wood}, \mathtt{grass}, \mathtt{sand}, \mathtt{dirt}$. 

\subsubsection{Optical Flow}
We estimate the optical flow using the implementation~\cite{2021mmflow} of an attention-based GMA model\cite{Jiang2021LearningTE}. We use the standard MMFlow configuration for the GMA pre-trained model on a mix of the datasets \cite{Jiang2021LearningTE, DFIB15, MIFDB16, Butler, Menze2018JPRS, sgteb}. While the model was pre-trained on images with size (768, 368), we resized the MOTChallenge images to (960, 540) at test time.

\section{Visual and Qualitative Results}
\label{supp:sec:visualizations}

This section shows a visual example of the difference between a BEV and a 2D image space prediction. Furthermore, we want to point to the additional scene videos also provided in the Supplementary material. 

\PAR{2D versus 3D.}
In~\Cref{fig:supp:2dvs3d} we show the trajectory prediction of our MG-GAN projected into the image (\Cref{fig:supp:2dvs3d_1}), the prediction in BEV (\Cref{fig:supp:2dvs3d_2}), and trajectory prediction in pixel space using a Kalman Filter (\Cref{fig:supp:2dvs3d_3}). We find the problem of the model reasoning in pixel space and cannot account for the effect of the camera perspective. As a consequence, this results in unrealistic motion in image space.

In contrast, we see in \Cref{fig:supp:2dvs3d_1} that our model predicting in BEV understands the spatial structure of the scene and is, therefore, able to predict the correct trajectory for the object and resolves the long-term occlusion.
\begin{figure}
    \centering
    \begin{subfigure}[t]{0.32\textwidth}
    \includegraphics[width=\textwidth]{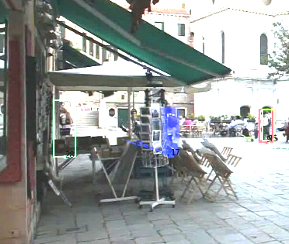}
    \caption{Projected MG-GAN prediction }
    \label{fig:supp:2dvs3d_1}
    \end{subfigure}
    \hfil
    \begin{subfigure}[t]{0.32\textwidth}
     \includegraphics[width=\textwidth]{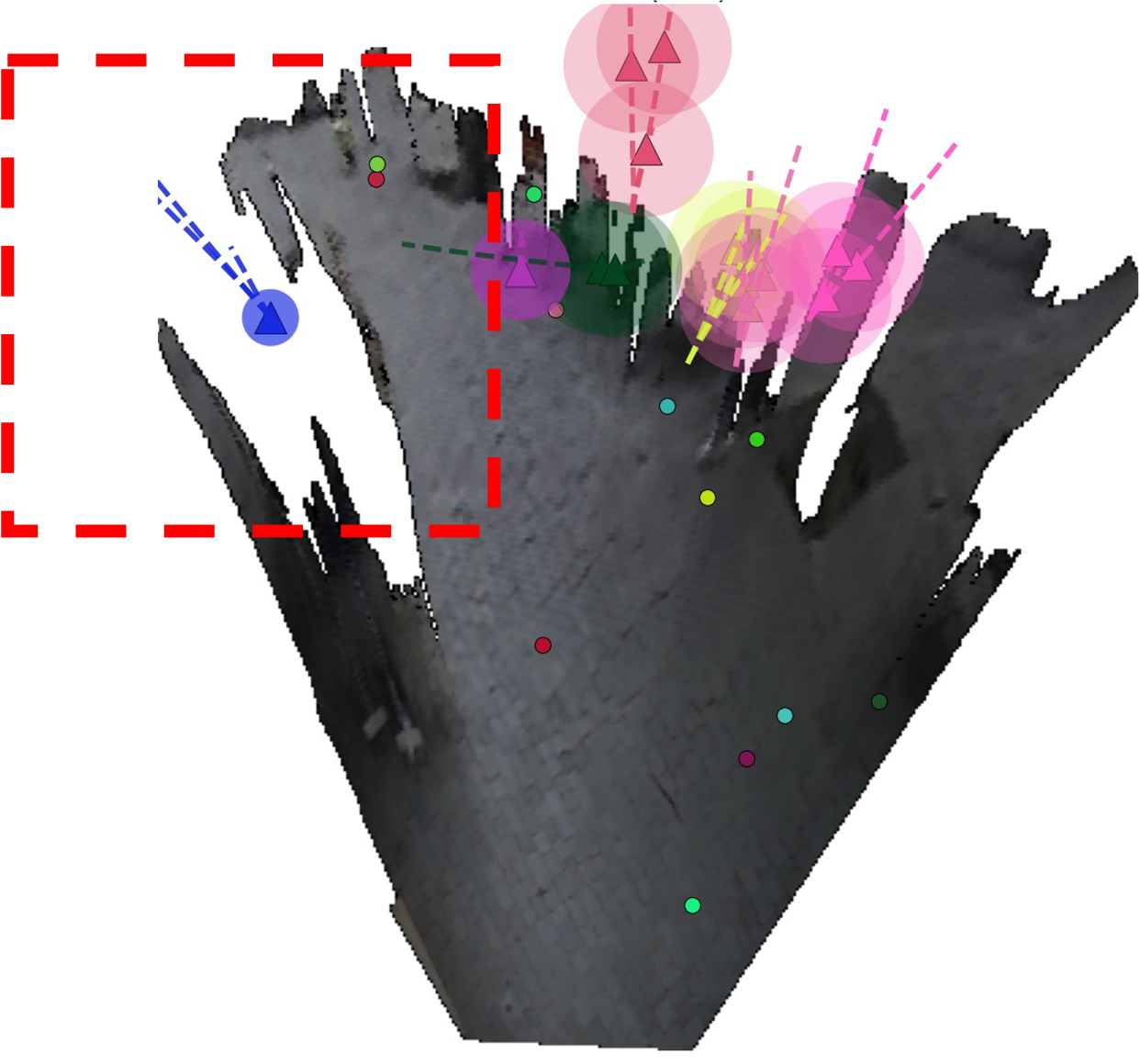}
    \caption{MG-GAN predictions in BEV}  
    \label{fig:supp:2dvs3d_2}
    \end{subfigure}
      \begin{subfigure}[t]{0.32\textwidth}
     \includegraphics[width=\textwidth]{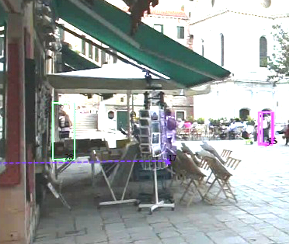}
    \caption{Prediction of Kalman Filter in Pixel Space}  \label{fig:supp:2dvs3d_3}
    \end{subfigure}
    \caption{Demonstration of prediction for MG-GAN in BEV and Kalman filter in pixel space.} 
    \label{fig:supp:2dvs3d}
\end{figure}

\PAR{Example Videos. }
In addition to the written supplementary material, we provide brief video clips of different MOT17 validation and test sequences with ByteTrack and Center Tracks. 

\begin{figure}

    \includegraphics[width=\textwidth]{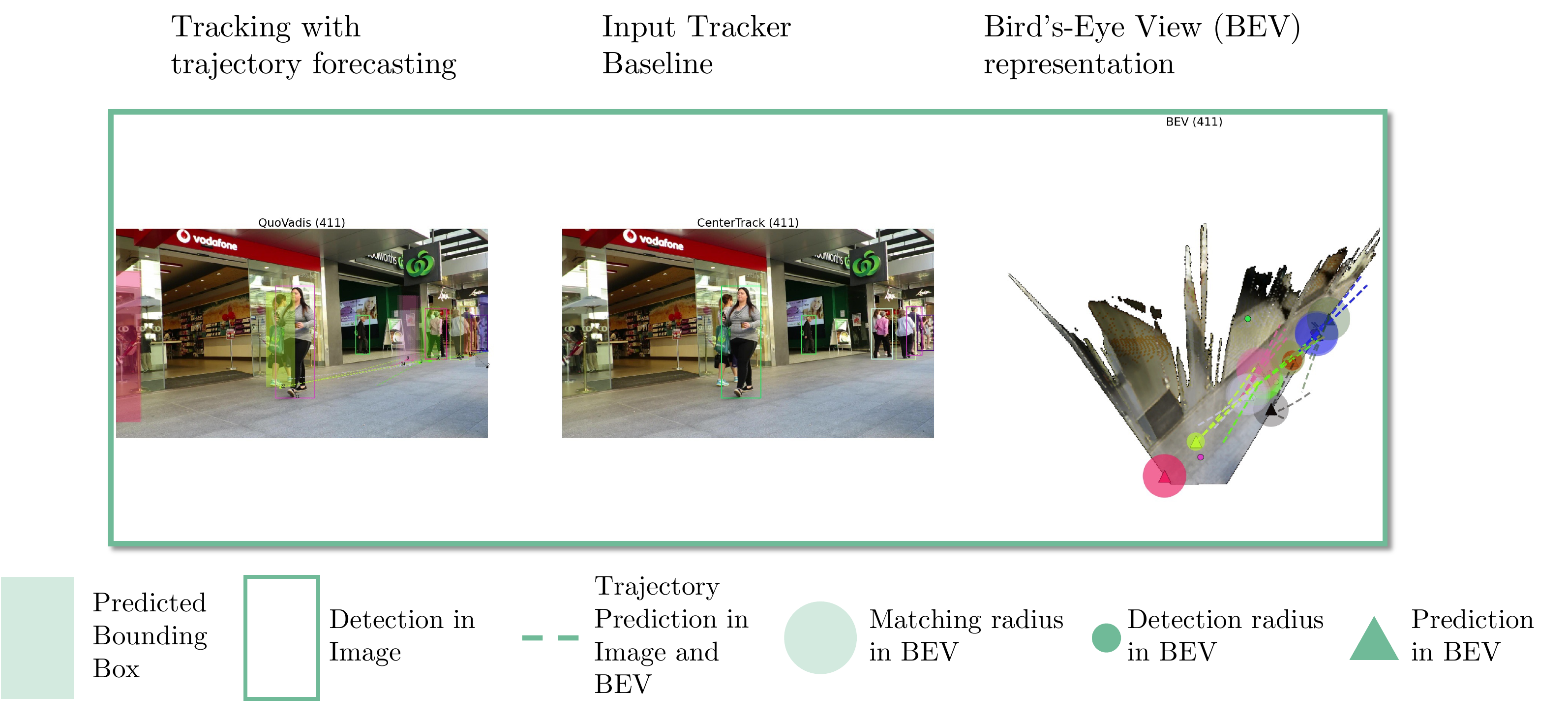}
    \caption{Description of supplementary sequence videos.}
    \label{fig:supp:video}

\end{figure}

In~\Cref{fig:supp:video}, we give a brief description of the format of the provided video sequences.
We show our predictor output on the left side, the baseline tracker output in the middle, and the online BEV prediction and reasoning on the right side. For our model, we show the tracker detection and predictions in BEV, including their projection in the image.
\section{Information on computation of ID Recall}
In the introduction, we present the performance of the baseline trackers compared to our trajectory forecasting model on how well they can re-associate tracks after occlusion from occlusions. We measure the performance as a fraction of ground-truth tracks detected and assigned correctly before and after occlusion.  

As the first step, we need to identify the ground-truth occlusion regions for every sequence. We use the visibility scores of objects and threshold those into a binary visibility flag, stating whether an object is visible in a given frame. Then, we apply a minimum rolling window on the visibility flags to get connected components and to smooth the deviations of the visibility flag values. The rolling window also includes visible frames before and after the occlusion, where the actual IDSW may happen. We compute the frame ids where an occlusion starts and ends by extracting all connected components, with the visibility flag being $0$. We only consider components where the object is visible before and after the occlusion.

Finally, we check for every tracker if the tracker detected an object at the start and the end of the occlusion component and if the track ids between the beginning and start match. We use $\tau_\text{vis} = 0.1$ as visibility threshold and a temporal window size $\mathtt{ws} = 5$ as hyperparameters.

%% file: main.bbl
\begin{thebibliography}{10}

\bibitem{Alahi16CVPR}
Alexandre Alahi, Kratarth Goel, Vignesh Ramanathan, Alexandre Robicquet,
  Li~Fei-Fei, and Silvio Savarese.
\newblock {Social LSTM}: Human trajectory prediction in crowded spaces.
\newblock In {\em Conference on Computer Vision and Pattern Recognition}, 2016.

\bibitem{social_ways}
Javad Amirian, Jean-Bernard Hayet, and Julien Pettr{\'e}.
\newblock {Social Ways: Learning multi-modal distributions of pedestrian
  trajectories with {GAN}s}.
\newblock In {\em Conference on Computer Vision and Pattern Recognition
  Workshops}, 2019.

\bibitem{anjum2020ctmc}
Samreen Anjum and Danna Gurari.
\newblock {CTMC}: Cell tracking with mitosis detection dataset challenge.
\newblock In {\em Conference on Computer Vision and Pattern Recognition
  Workshops}, 2020.

\bibitem{Bergmann19ICCV}
Philipp Bergmann, Tim Meinhardt, and Laura Leal-Taix{\'e}.
\newblock Tracking without bells and whistles.
\newblock In {\em International Conference on Computer Vision}, 2019.

\bibitem{bewley16icip}
Alex Bewley, Zongyuan Ge, Lionel Ott, Fabio Ramos, and Ben Upcroft.
\newblock Simple online and realtime tracking.
\newblock In {\em International Conference on Image Processing}, 2016.

\bibitem{bhat2020adabins}
Shariq~Farooq Bhat, Ibraheem Alhashim, and Peter Wonka.
\newblock Adabins: Depth estimation using adaptive bins.
\newblock In {\em Conference on Computer Vision and Pattern Recognition}, 2021.

\bibitem{best-of-many-sampling}
Apratim Bhattacharyya, Bernt Schiele, and Mario Fritz.
\newblock Accurate and diverse sampling of sequences based on a “{Best of
  Many}” sample objective.
\newblock In {\em Conference on Computer Vision and Pattern Recognition}, 2018.

\bibitem{Braso20CVPR}
Guillem Braso and Laura Leal-Taixe.
\newblock Learning a neural solver for multiple object tracking.
\newblock In {\em Conference on Computer Vision and Pattern Recognition}, 2020.

\bibitem{Butler}
D.~J. Butler, J.~Wulff, G.~B. Stanley, and M.~J. Black.
\newblock A naturalistic open source movie for optical flow evaluation.
\newblock In {\em European Conference on Computer Vision}, 2012.

\bibitem{Caesar20CVPR}
Holger Caesar, Varun Bankiti, Alex~H. Lang, Sourabh Vora, Venice~Erin Liong,
  Qiang Xu, Anush Krishnan, Yu~Pan, Giancarlo Baldan, and Oscar Beijbom.
\newblock {nuScenes}: A multimodal dataset for autonomous driving.
\newblock In {\em Conference on Computer Vision and Pattern Recognition}, 2020.

\bibitem{Choi15ICCV}
Wongun Choi.
\newblock Near-online multi-target tracking with aggregated local flow
  descriptor.
\newblock In {\em International Conference on Computer Vision}, 2015.

\bibitem{2021mmflow}
MMFlow Contributors.
\newblock {MMFlow}: Openmmlab optical flow toolbox and benchmark.
\newblock \url{https://github.com/open-mmlab/mmflow}, 2021.

\bibitem{dendorfer21iccv}
Patrick Dendorfer, Sven Elflein, and Laura Leal-Taix{\'e}.
\newblock {MG-GAN}: A multi-generator model preventing out-of-distribution
  samples in pedestrian trajectory prediction.
\newblock In {\em International Conference on Computer Vision}, 2021.

\bibitem{dendorfer20accv}
Patrick Dendorfer, Aljo\v{s}a O\v{s}ep, and Laura Leal-Taix\'e.
\newblock {Goal-{GAN}: Multimodal Trajectory Prediction Based on Goal Position
  Estimation}.
\newblock In {\em Asian Conference on Computer Vision}, 2020.

\bibitem{dendorfer20ijcv}
Patrick Dendorfer, Aljo\v{s}a O\v{s}ep, Anton Milan, Konrad Schindler, Daniel
  Cremers, Ian Reid, Stefan Roth, and Laura Leal-Taix{\'e}.
\newblock {MOTChallenge}: A benchmark for single-camera multiple target
  tracking.
\newblock In {\em International Journal of Computer Vision}, 2020.

\bibitem{MOT20}
Patrick Dendorfer, Hamid Rezatofighi, Anton Milan, Javen Shi, Daniel Cremers,
  Ian Reid, Stefan Roth, Konrad Schindler, and Laura Leal-Taix\'{e}.
\newblock {MOT20:} {A} benchmark for multi object tracking in crowded scenes.
\newblock In {\em arXiv preprint arXiv:2003.09003}, 2020.

\bibitem{DFIB15}
A.~Dosovitskiy, P.~Fischer, E.~Ilg, P.~H{\"a}usser, C.~Haz{\i}rba{\c{s}},
  V.~Golkov, P.~v.d. Smagt, D.~Cremers, and T.~Brox.
\newblock Flownet: Learning optical flow with convolutional networks.
\newblock In {\em International Conference on Computer Vision}, 2015.

\bibitem{dosovitskiy2021image}
Alexey Dosovitskiy, Lucas Beyer, Alexander Kolesnikov, Dirk Weissenborn,
  Xiaohua Zhai, Thomas Unterthiner, Mostafa Dehghani, Matthias Minderer, Georg
  Heigold, Sylvain Gelly, Jakob Uszkoreit, and Neil Houlsby.
\newblock An image is worth 16x16 words: Transformers for image recognition at
  scale.
\newblock In {\em International Conference on Learning Representations}, 2021.

\bibitem{fabbri2021motsynth}
Matteo Fabbri, Guillem Bras{\'o}, Gianluca Maugeri, Orcun Cetintas, Riccardo
  Gasparini, Aljo{\v{s}}a O{\v{s}}ep, Simone Calderara, Laura Leal-Taix{\'e},
  and Rita Cucchiara.
\newblock {MOTSynth}: How can synthetic data help pedestrian detection and
  tracking?
\newblock In {\em International Conference on Computer Vision}, 2021.

\bibitem{Geiger14TPAMI}
Andreas Geiger, Martin Lauer, Christian Wojek, Christoph Stiller, and Raquel
  Urtasun.
\newblock {3D} traffic scene understanding from movable platforms.
\newblock In {\em Transactions on Pattern Analysis and Machine Intelligence},
  2014.

\bibitem{Geiger12CVPR}
Andreas Geiger, Philip Lenz, and Raquel Urtasun.
\newblock Are we ready for autonomous driving? {T}he {KITTI} vision benchmark
  suite.
\newblock In {\em Conference on Computer Vision and Pattern Recognition}, 2012.

\bibitem{nips04gan}
Ian Goodfellow, Jean Pouget-Abadie, Mehdi Mirza, Bing Xu, David Warde-Farley,
  Sherjil Ozair, Aaron Courville, and Yoshua Bengio.
\newblock Generative adversarial nets.
\newblock In {\em Conference on Neural Information Processing Systems}, 2014.

\bibitem{Gupta18CVPR}
Agrim Gupta, Justin Johnson, Li~Fei-Fei, Silvio Savarese, and Alexandre Alahi.
\newblock Social gan: Socially acceptable trajectories with generative
  adversarial networks.
\newblock In {\em Conference on Computer Vision and Pattern Recognition}, 2018.

\bibitem{He16CVPR}
K.~He, X.~Zhang, S.~Ren, and J.~Sun.
\newblock Deep residual learning for image recognition.
\newblock In {\em Conference on Computer Vision and Pattern Recognition}, 2016.

\bibitem{social_force}
Dirk Helbing and Péter Molnár.
\newblock {Social force model for pedestrian dynamics}.
\newblock In {\em Physical Review E}, 1995.

\bibitem{Hu18arxiv}
Hou-Ning Hu, Qi-Zhi Cai, Dequan Wang, Ji~Lin, Min Sun, Philipp Krahenbuhl,
  Trevor Darrell, and Fisher Yu.
\newblock {Joint Monocular 3D Vehicle Detection and Tracking}.
\newblock In {\em International Conference on Computer Vision}, 2018.

\bibitem{Huang08ECCV}
Chang Huang, Bo~Wu, and Ramakant Nevatia.
\newblock Robust object tracking by hierarchical association of detection
  responses.
\newblock In {\em European Conference on Computer Vision}, 2008.

\bibitem{Jiang2021LearningTE}
Shihao Jiang, Dylan Campbell, Yao Lu, Hongdong Li, and Richard~I. Hartley.
\newblock Learning to estimate hidden motions with global motion aggregation.
\newblock In {\em International Conference on Computer Vision}, 2021.

\bibitem{khurana2021detecting}
Tarasha Khurana, Achal Dave, and Deva Ramanan.
\newblock Detecting invisible people.
\newblock In {\em International Conference on Computer Vision}, 2021.

\bibitem{sgteb}
Daniel Kondermann, Rahul Nair, Stephan Meister, Wolfgang Mischler, Burkhard
  Güssefeld, Katrin Honauer, Sabine Hofmann, Claus Brenner, and Bernd Jähne.
\newblock Stereo ground truth with error bars.
\newblock In {\em Asian Conference on Computer Vision}, 2014.

\bibitem{bigat}
Vineet Kosaraju, Amir Sadeghian, Roberto Mart{\'\i}n-Mart{\'\i}n, Ian Reid,
  Hamid Rezatofighi, and Silvio Savarese.
\newblock {Social-BiGAT: Multimodal trajectory forecasting using {Bicycle-GAN}
  and graph attention networks}.
\newblock In {\em Conference on Neural Information Processing Systems}, 2019.

\bibitem{LealTaixe16CVPRW}
Laura Leal-Taix{\'e}, Cristian Canton-Ferrer, and Konrad Schindler.
\newblock Learning by tracking: Siamese cnn for robust target association.
\newblock In {\em Conference on Computer Vision and Pattern Recognition
  Workshops}, 2016.

\bibitem{leal14cvpr}
Laura Leal-Taix{\'e}, Michele Fenzi, Alina Kuznetsova, Bodo Rosenhahn, and
  Silvio Savarese.
\newblock Learning an image-based motion context for multiple people tracking.
\newblock In {\em Conference on Computer Vision and Pattern Recognition}, 2014.

\bibitem{leibe08TPAMI}
Bastian Leibe, Konrad Schindler, Nico Cornelis, and Luc~Van Gool.
\newblock Coupled object detection and tracking from static cameras and moving
  vehicles.
\newblock In {\em Transactions on Pattern Analysis and Machine Intelligence},
  2008.

\bibitem{UCY-data}
Alon Lerner, Yiorgos Chrysanthou, and Dani Lischinski.
\newblock {Crowds by Example}.
\newblock In {\em Comput. Graph. Forum}, 2007.

\bibitem{Li09CVPR}
Yuan Li, Chang Huang, and Ramkat Nevatia.
\newblock Learning to associate: Hybridboosted multi-target tracker for crowded
  scene.
\newblock In {\em Conference on Computer Vision and Pattern Recognition}, 2009.

\bibitem{Zhuang20ECCV}
Chao Liang, Zhipeng Zhang, Xue Zhou, Bing Li, Shuyuan Zhu, and Weiming Hu.
\newblock Rethinking the competition between detection and r{ReID} in
  multiobject tracking.
\newblock In {\em Transactions on Image Processing}, 2022.

\bibitem{liang2020garden}
Junwei Liang, Lu~Jiang, Kevin Murphy, Ting Yu, and Alexander Hauptmann.
\newblock {The Garden of Forking Paths: Towards Multi-Future Trajectory
  Prediction}.
\newblock In {\em Conference on Computer Vision and Pattern Recognition}, 2020.

\bibitem{Liu20ijcai}
Qiankun Liu, Qi~Chu, Bin Liu, and Nenghai Yu.
\newblock {GSM}: Graph similarity model for multi-object tracking.
\newblock In {\em International Joint Conferences on Artificial Intelligence},
  2020.

\bibitem{loshchilov2019decoupled}
Ilya Loshchilov and Frank Hutter.
\newblock Decoupled weight decay regularization.
\newblock In {\em International Conference on Learning Representations}, 2019.

\bibitem{luiten20ijcv}
Jonathon Luiten, Aljo\v{s}a O\v{s}ep, Patrick Dendorfer, Philip Torr, Andreas
  Geiger, Laura Leal-Taix{\'e}, and Bastian Leibe.
\newblock {HOTA}: A higher order metric for evaluating multi-object tracking.
\newblock In {\em International Journal of Computer Vision}, 2020.

\bibitem{Mangalam_2021_ICCV}
Karttikeya Mangalam, Yang An, Harshayu Girase, and Jitendra Malik.
\newblock From goals, waypoints \& paths to long term human trajectory
  forecasting.
\newblock In {\em International Conference on Computer Vision}, 2021.

\bibitem{mangalam2020pecnet}
Karttikeya Mangalam, Harshayu Girase, Shreyas Agarwal, Kuan-Hui Lee, Ehsan
  Adeli, Jitendra Malik, and Adrien Gaidon.
\newblock It is not the journey but the destination: Endpoint conditioned
  trajectory prediction.
\newblock In {\em European Conference on Computer Vision}, 2020.

\bibitem{martin2021jrdb}
Roberto Martin-Martin, Mihir Patel, Hamid Rezatofighi, Abhijeet Shenoi,
  JunYoung Gwak, Eric Frankel, Amir Sadeghian, and Silvio Savarese.
\newblock {JRDB}: A dataset and benchmark of egocentric robot visual perception
  of humans in built environments.
\newblock In {\em Transactions on Pattern Analysis and Machine Intelligence},
  2021.

\bibitem{MIFDB16}
N.~Mayer, E.~Ilg, P.~H{\"a}usser, P.~Fischer, D.~Cremers, A.~Dosovitskiy, and
  T.~Brox.
\newblock A large dataset to train convolutional networks for disparity,
  optical flow, and scene flow estimation.
\newblock In {\em Conference on Computer Vision and Pattern Recognition}, 2016.

\bibitem{Menze2018JPRS}
Moritz Menze, Christian Heipke, and Andreas Geiger.
\newblock Object scene flow.
\newblock In {\em Journal of Photogrammetry and Remote Sensing}, 2018.

\bibitem{MOT17}
A.~Milan, L.~Leal-Taix\'{e}, I.~Reid, S.~Roth, and K.~Schindler.
\newblock {MOT}16: {A} benchmark for multi-object tracking.
\newblock In {\em arXiv preprint arXiv:1603.00831}, 2016.

\bibitem{Milan14TPAMI}
Anton Milan, Stefan Roth, and Konrad Schindler.
\newblock Continuous energy minimization for multitarget tracking.
\newblock In {\em Transactions on Pattern Analysis and Machine Intelligence},
  2014.

\bibitem{Milan13CVPR}
Anton Milan, Konrad Schindler, and Stefan Roth.
\newblock Detection- and trajectory-level exclusion in multiple object
  tracking.
\newblock In {\em Conference on Computer Vision and Pattern Recognition}, 2013.

\bibitem{Osep17ICRA}
Aljo\v{s}a O\v{s}ep, Wolfgang Mehner, Markus Mathias, and Bastian Leibe.
\newblock Combined image- and world-space tracking in traffic scenes.
\newblock In {\em International Conference on Robotics and Automation}, 2017.

\bibitem{Pang20CVPR}
Jiangmiao Pang, Linlu Qiu, Xia Li, Haofeng Chen, Qi~Li, Trevor Darrell, and
  Fisher Yu.
\newblock Quasi-dense similarity learning for multiple object tracking.
\newblock In {\em Conference on Computer Vision and Pattern Recognition}, 2020.

\bibitem{pedersen20203d}
Malte Pedersen, Joakim~Bruslund Haurum, Stefan~Hein Bengtson, and Thomas~B
  Moeslund.
\newblock {3D-ZeF: A 3D zebrafish tracking benchmark dataset}.
\newblock In {\em Conference on Computer Vision and Pattern Recognition}, 2020.

\bibitem{ETH-data}
S.~Pellegrini, Andreas Ess, and L.~Gool.
\newblock Improving data association by joint modeling of pedestrian
  trajectories and groupings.
\newblock In {\em European Conference on Computer Vision}, 2010.

\bibitem{Reid79TAC}
Donald~B Reid.
\newblock An algorithm for tracking multiple targets.
\newblock In {\em Transactions on Automatic Control}, 1979.

\bibitem{stanforddronedataset}
Alexandre Robicquet, Amir Sadeghian, Alexandre Alahi, and Silvio Savarese.
\newblock {Learning social etiquette: Human trajectory understanding in crowded
  scenes}.
\newblock In {\em European Conference on Computer Vision}, 2016.

\bibitem{sadeghian2018sophie}
Amir Sadeghian, Vineet Kosaraju, Ali Sadeghian, Noriaki Hirose, Hamid
  Rezatofighi, and Silvio Savarese.
\newblock {Sophie: An attentive {GAN} for predicting paths compliant to social
  and physical constraints}.
\newblock In {\em Conference on Computer Vision and Pattern Recognition}, 2019.

\bibitem{smith2018superconvergence}
Leslie~N. Smith and Nicholay Topin.
\newblock Super-convergence: Very fast training of neural networks using large
  learning rates.
\newblock In {\em arXiv preprint arXiv:1708.07120}, 2018.

\bibitem{smith18batchsize}
Samuel~L. Smith, Pieter-Jan Kindermans, Chris Ying, and Quoc~V. Le.
\newblock Don't decay the learning rate, increase the batch size.
\newblock In {\em International Conference on Learning Representations}, 2018.

\bibitem{Son17CVPR}
Jeany Son, Mooyeol Baek, Minsu Cho, and Bohyung Han.
\newblock Multi-object tracking with quadruplet convolutional neural networks.
\newblock In {\em Conference on Computer Vision and Pattern Recognition}, 2017.

\bibitem{Stadler_2022_WACV}
Daniel Stadler and J\"urgen Beyerer.
\newblock Modelling ambiguous assignments for multi-person tracking in crowds.
\newblock In {\em Winter Conference on Applications of Computer Vision}, 2022.

\bibitem{9663836}
Daniel Stadler and Jürgen Beyerer.
\newblock On the performance of crowd-specific detectors in multi-pedestrian
  tracking.
\newblock In {\em International Conference on Advanced Video and Signal-Based
  Surveillance}, 2021.

\bibitem{sun20CVPR}
Pei Sun, Henrik Kretzschmar, Xerxes Dotiwalla, Aurelien Chouard, Vijaysai
  Patnaik, Paul Tsui, James Guo, Yin Zhou, Yuning Chai, Benjamin Caine, et~al.
\newblock Scalability in perception for autonomous driving: Waymo open dataset.
\newblock In {\em Conference on Computer Vision and Pattern Recognition}, 2020.

\bibitem{Sun20transtrack}
Peize Sun, Jinkun Cao, Yi~Jiang, Rufeng Zhang, Enze Xie, Zehuan Yuan, Changhu
  Wang, and Ping Luo.
\newblock Transtrack: Multiple-object tracking with transformer.
\newblock In {\em arXiv preprint arXiv: 2012.15460}, 2020.

\bibitem{GANmanifold}
Ugo Tanielian, Thibaut Issenhuth, Elvis Dohmatob, and Jeremie Mary.
\newblock Learning disconnected manifolds: a no gans land.
\newblock In {\em Proceedings of Machine Learning and Systems}, 2020.

\bibitem{tokmakov2021learning}
Pavel Tokmakov, Jie Li, Wolfram Burgard, and Adrien Gaidon.
\newblock Learning to track with object permanence.
\newblock In {\em Conference on Computer Vision and Pattern Recognition}, 2021.

\bibitem{Voigtlaender19CVPR}
Paul Voigtlaender, Michael Krause, Aljosa Osep, Jonathon Luiten, B.B.G Sekar,
  Andreas Geiger, and Bastian Leibe.
\newblock {MOTS}: Multi-object tracking and segmentation.
\newblock In {\em Conference on Computer Vision and Pattern Recognition}, 2019.

\bibitem{wang:cvpr:21}
Qiang Wang, Yun Zheng, Pan Pan, and Yinghui Xu.
\newblock Multiple object tracking with correlation learning.
\newblock In {\em Conference on Computer Vision and Pattern Recognition}, 2021.

\bibitem{Wang20ECCV}
Zhongdao Wang, Liang Zheng, Yixuan Liu, Yali Li, and Shengjin Wang.
\newblock Towards real-time multi-object tracking.
\newblock In {\em European Conference on Computer Vision}, 2020.

\bibitem{Wen15arxiv}
Longyin Wen, Dawei Du, Zhaowei Cai, Zhen Lei, Ming-Ching Chang, Honggang Qi,
  Jongwoo Lim, Ming-Hsuan Yang, and Siwei Lyu.
\newblock {UA-DETRAC: A new benchmark and protocol for multi-object detection
  and tracking}.
\newblock In {\em Computer Vision and Image Understanding}, 2015.

\bibitem{weng20CVPR}
Xinshuo Weng, Yongxin Wang, Yunze Man, and Kris Kitani.
\newblock {GNN3DMOT: Graph Neural Network for 3D Multi-Object Tracking with
  Multi-Feature Learning}.
\newblock In {\em Conference on Computer Vision and Pattern Recognition}, 2020.

\bibitem{Wu09IJCV}
Bo~Wu and Ramkat Nevatia.
\newblock Detection and segmentation of multiple, partially occluded objects by
  grouping, merging, assigning part detection responses.
\newblock In {\em International Journal of Computer Vision}, 2009.

\bibitem{Wu21CVPR}
Jialian Wu, Jiale Cao, Liangchen Song, Yu~Wang, Ming Yang, and Junsong Yuan.
\newblock Track to detect and segment: An online multi-object tracker.
\newblock In {\em Conference on Computer Vision and Pattern Recognition}, 2021.

\bibitem{wu2019detectron2}
Yuxin Wu, Alexander Kirillov, Francisco Massa, Wan-Yen Lo, and Ross Girshick.
\newblock Detectron2.
\newblock \url{https://github.com/facebookresearch/detectron2}, 2019.

\bibitem{xu2021transcenter}
Yihong Xu, Yutong Ban, Guillaume Delorme, Chuang Gan, Daniela Rus, and Xavier
  Alameda-Pineda.
\newblock Transcenter: Transformers with dense queries for multiple-object
  tracking.
\newblock In {\em arXiv preprint arXiv:2103.15145}, 2021.

\bibitem{xu20cvpr}
Yihong Xu, Aljo\v{s}a O\v{s}ep, Yutong Ban, Radu Horaud, Laura Leal-Taix{\'e},
  and Xavier Alameda-Pineda.
\newblock How to train your deep multi-object tracker.
\newblock In {\em Conference on Computer Vision and Pattern Recognition}, 2020.

\bibitem{re_mot}
Fan Yang, Xin Chang, Sakriani Sakti, Yang Wu, and Satoshi Nakamura.
\newblock {ReMOT}: A model-agnostic refinement for multiple object tracking.
\newblock In {\em Image and Vision Computing}, 2020.

\bibitem{bdd100k}
Fisher Yu, Haofeng Chen, Xin Wang, Wenqi Xian, Yingying Chen, Fangchen Liu,
  Vashisht Madhavan, and Trevor Darrell.
\newblock Bdd100k: A diverse driving dataset for heterogeneous multitask
  learning.
\newblock In {\em Conference on Computer Vision and Pattern Recognition}, 2020.

\bibitem{zaech2022learnable}
Jan-Nico Zaech, Alexander Liniger, Dengxin Dai, Martin Danelljan, and Luc
  Van~Gool.
\newblock {Learnable Online Graph Representations for 3D Multi-Object
  Tracking}.
\newblock In {\em Robotics and Automation Society}, 2022.

\bibitem{Zhang08CVPR}
Li~Zhang, Li~Yuan, and Ramakant Nevatia.
\newblock Global data association for multi-object tracking using network
  flows.
\newblock In {\em Conference on Computer Vision and Pattern Recognition}, 2008.

\bibitem{byte_track}
Yifu Zhang, Peize Sun, Yi~Jiang, Dongdong Yu, Fucheng Weng, Zehuan Yuan, Ping
  Luo, Wenyu Liu, and Xinggang Wang.
\newblock {ByteTrack}: Multi-object tracking by associating every detection
  box.
\newblock In {\em European Conference on Computer Vision}, 2022.

\bibitem{Zhang2021IJCV}
Yifu Zhang, Chunyu Wang, Xinggang Wang, Wenjun Zeng, and Wenyu Liu.
\newblock {FairMOT}: On the fairness of detection and re-identification in
  multiple object tracking.
\newblock In {\em International Journal of Computer Vision}, 2021.

\bibitem{zhou2020tracking}
Xingyi Zhou, Vladlen Koltun, and Philipp Kr{\"a}henb{\"u}hl.
\newblock Tracking objects as points.
\newblock In {\em European Conference on Computer Vision}, 2020.

\end{thebibliography}
